\pdfoutput=1

\documentclass[11pt]{article}

\usepackage[preprint]{acl}

\usepackage{times}
\usepackage{latexsym}

\usepackage[T1]{fontenc}

\usepackage{inconsolata}

\usepackage[utf8]{inputenc} 

\usepackage{hyperref}       
\usepackage{url}            
\usepackage{booktabs}       
\usepackage{amsfonts}       
\usepackage{nicefrac}       
\usepackage{microtype}      

\usepackage{multirow}
\usepackage{amsmath}
\usepackage{graphicx}
\usepackage{svg}
\usepackage{cleveref}
\usepackage{scalefnt}
\usepackage{chngcntr}
\usepackage{amssymb}
\usepackage{pifont}
\usepackage{titletoc} 
\usepackage{wrapfig}
\usepackage{caption}

\newcommand{\xmark}{\ding{55}}
\newcommand{\cross}{\textcolor{red}{\xmark}}
\colorlet{Mycolor1}{green!100}
\newcommand{\cmark}{\textcolor{Mycolor1}{\ding{51}}}

\newcommand{\methodname}{AutoGUI}

\definecolor{DarkGreen}{RGB}{8,120,48}
\newcommand{\gain}[1]{\scalebox{0.8}{\textcolor{DarkGreen}{#1}}}

\title{AutoGUI: Scaling GUI Grounding with Automatic Functionality Annotations from LLMs}

\author{
 \textbf{Hongxin Li\textsuperscript{*1,2,3}},
 \textbf{Jingfan Chen\textsuperscript{*5}},
 \textbf{Jingran Su\textsuperscript{*5}},
 \textbf{Yuntao Chen$^{\dagger}$\textsuperscript{4}}
\\
 \textbf{Qing Li\textsuperscript{5}},
 \textbf{Zhaoxiang Zhang$^{\dagger}$\textsuperscript{1,2,3,6}}
\\
 \textsuperscript{1}University of Chinese Academy of Sciences (UCAS)\\
 \textsuperscript{2}New Laboratory of Pattern Recognition (NLPR), CASIA\\
 \textsuperscript{3}State Key Laboratory of Multimodal Artificial Intelligence Systems (MAIS), CASIA\\
 \textsuperscript{4}Hong Kong Institute of Science \& Innovation, CASIA\\
 \textsuperscript{5}The Hong Kong Polytechnic University \textsuperscript{6}Shanghai Artificial Intelligence Laboratory \\
  \\
}

\begin{document}
\maketitle

\newcommand{\myfootnote}[1]{%
  \begingroup
  \renewcommand\thefootnote{}\footnotetext{#1}%
  \endgroup
}

\myfootnote{$^{*}$ Equal contribution.}
\myfootnote{$^{\dagger}$ Equally advising corresponding authors. E-mails: zhaoxiang.zhang@ia.ac.cn, chenyuntao08@gmail.com}

\begin{abstract}
User interface understanding with vision-language models (VLMs) has received much attention due to its potential for enhancing software automation.
However, existing datasets used to build UI-VLMs either only contain large-scale context-free element annotations or contextualized functional descriptions for elements at a small scale.
In this work, we propose the \textbf{AutoGUI} pipeline for automatically annotating UI elements with detailed functionality descriptions at scale.
Specifically, we leverage large language models (LLMs) to infer element functionality by comparing UI state changes before and after simulated interactions. To improve annotation quality, we propose LLM-aided rejection and verification, eliminating invalid annotations without human labor.
We construct a high-quality AutoGUI-704k dataset using the proposed pipeline, featuring diverse and detailed functionality annotations that are hardly provided by previous datasets.
Human evaluation shows that we achieve annotation correctness comparable to a trained human annotator. Extensive experiments show that our dataset remarkably enhances VLM's UI grounding capabilities and exhibits significant scaling effects. We also show the interesting potential use of our dataset in UI agent tasks. Please view our project at \url{https://autogui-project.github.io/}.
\end{abstract}

\section{Introduction}

User interface understanding with visual language models(VLMs)~\citep{hong2023cogagent,you2024ferret,osatlas} has received wide attention due to its potential in fundamentally transforming how we interact with software~\citep{OSWorld}.

While recent work has made progress by employing structural mapping between UI code and visual layout, such as UI REG/REC\citep{hong2023cogagent,Li2020WidgetCG} and layout-to-code conversion~\citep{xia2024chartx, liu2023deplot,baechler2024screenai}, a more critical challenge remains: understanding the semantic purpose and interactive affordance of individual UI elements, known as \textit{functionality understanding}.

\begin{figure}[t]
    \centering
    \includegraphics[width=1.0\linewidth]{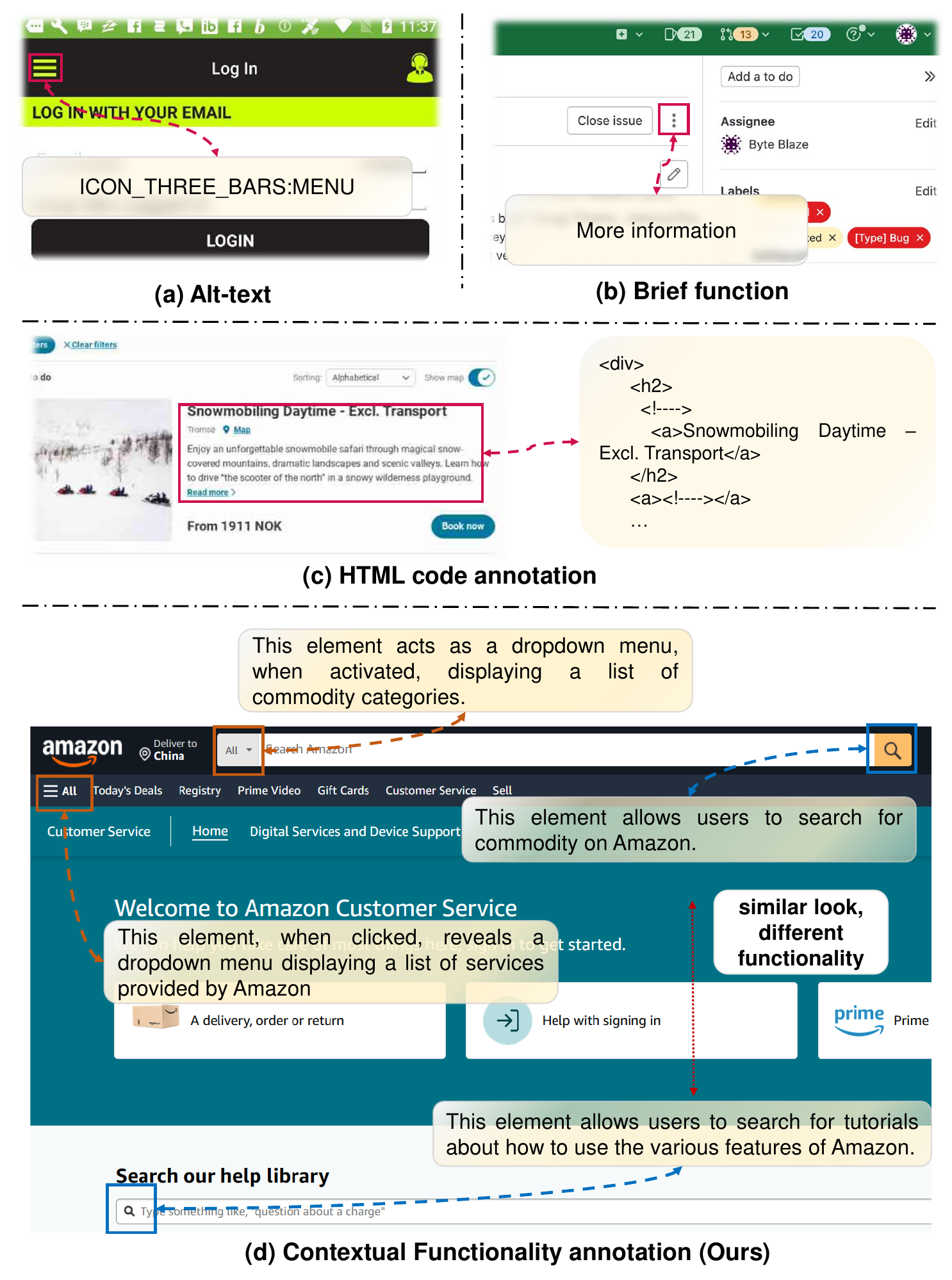}
    \caption{Our annotations are rich in functional semantics (bottom) compared with existing UI datasets.}
    \label{fig: functionality vs others}
\end{figure}

Accurate functionality understanding requires VLMs to possess strong element grounding capabilities - the ability to connect fine-grained visual elements with their referring expressions. To enhance this capability, large-scale training data is indispensable.
However, the scale of open-source datasets with detailed element annotations~\citep{Li2020WidgetCG,Li2020MappingNL,kapoor2024omniact,uground} is unsatisfactory, significantly smaller than natural image datasets such as LAION-5B~\citep{LAION5B}.
Additionally, traditional annotation methods~\citep{Deka2017RicoAM,Li2020WidgetCG} are labor-intensive, leading to prohibitive costs that hinder scalability.
Moreover, existing datasets typically focus on describing either element alt-texts~\citep{cheng2024seeclick}, or brief intents weakly related to UI context~\citep{Bai2021UIBertLG} shown in Fig.~\ref{fig: functionality vs others}.
These datasets lack contextual functional descriptions of UI elements, which poses a challenge for VLMs in comprehending the roles these elements serve within specific UI contexts, such as distinguishing between two visually similar magnifier icons that may represent distinct functionalities like searching and zooming.

To address the challenge, we propose \methodname{}, an automatic annotating pipeline that provides unlimited element functionality annotations. Our pipeline collects UI interaction trajectories and leverages large language models (LLMs) to infer element functionalities based on UI state changes, eliminating the need for manual annotation by human experts.
Initially, the proposed pipeline crawls a multitude of interaction trajectories on either a web browser or an Android emulator. Subsequently, we use open-source LLMs~\citep{llama3modelcard} to annotate the functionalities of elements on collected GUIs based on changes to UI contents when interacting with these elements. To ensure data quality, LLM-aided rejection is utilized to eliminate invalid samples, such as incompletely rendered UIs. Additionally, inspired by LLM verification~\citep{Weng2022LargeLM,Lightman2023LetsVS}, multiple LLMs are prompted as verifiers to identify false functionality descriptions. With both the rejection and verification processes, our pipeline removes unclear and invalid samples.

We curate the \methodname{}-704k dataset with the proposed pipeline, providing high-quality functionality grounding and referring tasks used to finetune and evaluate open-source VLMs.

Pioneer experiments find that our pipeline achieves annotation accuracy of \textbf{96.7\%} comparable to a trained human annotator.

Based on the collected \methodname{}-704k dataset, we finetune open-source VLMs and demonstrate that our data significantly enhances the VLMs' UI grounding accuracy and exhibits remarkable scaling effects. The results also show that our functionality annotation type is superior to the data type directly derived from web HTML code and metadata~\citep{hong2023cogagent,cheng2024seeclick}, serving as a promising data source for building VLMs capable of UI grounding. Moreover, VLMs trained with our data can assist in GUI agent tasks by refining element grounding, which shows more potential use of our dataset.
\section{Related Works}
\subsection{Recent Advancement of VLMs}
Recent research has enhanced LLMs with the capability of processing both visual and textual information~\citep{alayrac2022flamingo,liu2023llava,lin2023sphinx,chen2023internvl,lu2024deepseekvl,qwen2vl,wang2024visionllm,li2023monkey,zhang2024llamaadapter,you2024ferret,laurençon2024idefics,peng2024kosmos,driess2023palme}, opening the new field of VLM. Pioneering efforts Flamingo~\citep{alayrac2022flamingo} uses interleaved visual and language inputs as prompts and shows few-shot visual question-answering capability.

LLaVA~\citep{liu2023llava} and LLaMA-Adapter~\citep{zhang2024llamaadapter} have attempted to align vision encoders~\citep{dosovitskiy2021vit} with LLMs to enable visual instruction following. Advanced models such as InternVL~\citep{chen2023internvl} and Qwen2-VL series~\citep{qwen2vl} are further equipped with impressive high-resolution and multi-lingual understanding abilities. Additionally, VLM are applied to scenarios rich in textual imagery~\citep{2023-ureader,ye2023mplugdocowl,liu2024textmonkey} and embodied interactions~\citep{driess2023palme, kim24openvla}. Despite these advancements, VLMs lag in UI understanding probably due to data scarcity. This paper contributes an automatic UI annotation pipeline to tackle this challenge, aiming to expand the data available for training VLMs in this crucial area.

\begin{table*}[t]
\scriptsize
\centering
\caption{\textbf{Comparing our \methodname{} dataset with existing UI grounding datasets.} Multi-Res means the samples are collected on devices with various resolutions. Auto Anno. means the samples are collected autonomously. \#Anno. means the number of annotated samples provided by the datasets. (Methods combining open-source individual datasets are not compared.)}
\label{tab:data comparison}
\begin{tabular}{@{}ccccccc@{}}
\toprule
Dataset & UI Type & \begin{tabular}[c]{@{}c@{}}Multi\\ Res.\end{tabular} & \begin{tabular}[c]{@{}c@{}}Auto\\ Anno. \end{tabular} & \begin{tabular}[c]{@{}c@{}}Functionality\\ Annotation \\Type\end{tabular} & \#Anno. & Task \\ \midrule
S2W~\citep{Wang2021Screen2WordsAM} & Mobile & \cross  & \cross & N/A & 112k & Screen Summarization \\
Wid. Cap.~\citep{Li2020WidgetCG} & Mobile & \cross  & \cross & N/A & 163k & Element Captioning \\
RICOSCA~\citep{Li2020MappingNL} & Mobile & \cross  & \cross & N/A & 295k & Action Grounding \\
MoTIF~\citep{Burns2022ADF} & Mobile & \cross  & \cross & N/A & 6k & Mobile Navigation \\
RefExp~\citep{Bai2021UIBertLG} & Mobile & \cross  & \cross & N/A & 20.8k & Element Grounding \\
SeeClick Web~\citep{cheng2024seeclick} & Web & \cross  & \cmark & N/A & 271k & Element Grounding \\
MultiUI~\citep{hong2023cogagent} & Web, Mobile & \cmark  & \cmark &N/A & 3M & Act. \& Elem. Ground \\
UGround-Web~\citep{uground} & Web & \cmark  & \cmark & Brief & 1.3M & Element Grounding \\
UI REC/REG~\citep{hong2023cogagent} & Web & \cmark  & \cmark & N/A & 400k & Box2DOM, DOM2Box \\
Ferret-UI~\citep{you2024ferretui} & Mobile & \cmark  & \cmark & Brief & 250k & Elem. Ground \& Ref. \\
\textbf{\methodname{}} (ours) & Web, Mobile & \cmark & \cmark & \textbf{Contextual} & 704k & Functionality Ground \& Ref. \\ \bottomrule
\end{tabular}
\end{table*}

\subsection{Existing UI Datasets}

Unlike natural image datasets~\citep{Russakovsky2014ImageNetLS,LAION5B}, UI understanding datasets are much smaller. Early-stage datasets~\citep{Wang2021Screen2WordsAM,Li2020WidgetCG,Li2020MappingNL,Bai2021UIBertLG,Burns2022ADF} primarily annotate the RICO screenshot collection~\citep{deka2017rico}, which includes 72K screenshots from Android apps. Examples include Widget Captioning~\citep{Li2020WidgetCG}, which analyzes captions and linguistic features of UI elements, and RICOSCA~\citep{Li2020MappingNL}, which maps single-step instructions to UI locations. Recently, AITW~\citep{rawles2023android} and AndroidControl~\citep{androidcontrol} have been proposed to focus on interpreting high-level instructions in Android environments. To increase data scale, SeeClick~\citep{cheng2024seeclick}, CogAgent~\citep{hong2023cogagent}, and OS-ATLAS~\citep{osatlas} have utilized the UI metadata from Common Crawl webpages to produce massive element referring expressions. Several works~\cite{uground,showui,aguvis} also filter and combine existing datasets to produce all-in-one collections that incorporate diverse training tasks. In contrast, our \methodname{}-704k dataset contributes large-scale element functionality annotations, which convey contextual functionality semantics that are hardly provided in previous datasets. The advantages of our dataset are summarized in Tab.~\ref{tab:data comparison}.
\section{\methodname{}: Automatic Functionality Annotation Pipeline}
\label{sec: annotation pipeline}
\begin{figure*}[t]
    \centering
    \includegraphics[width=1.0\linewidth]{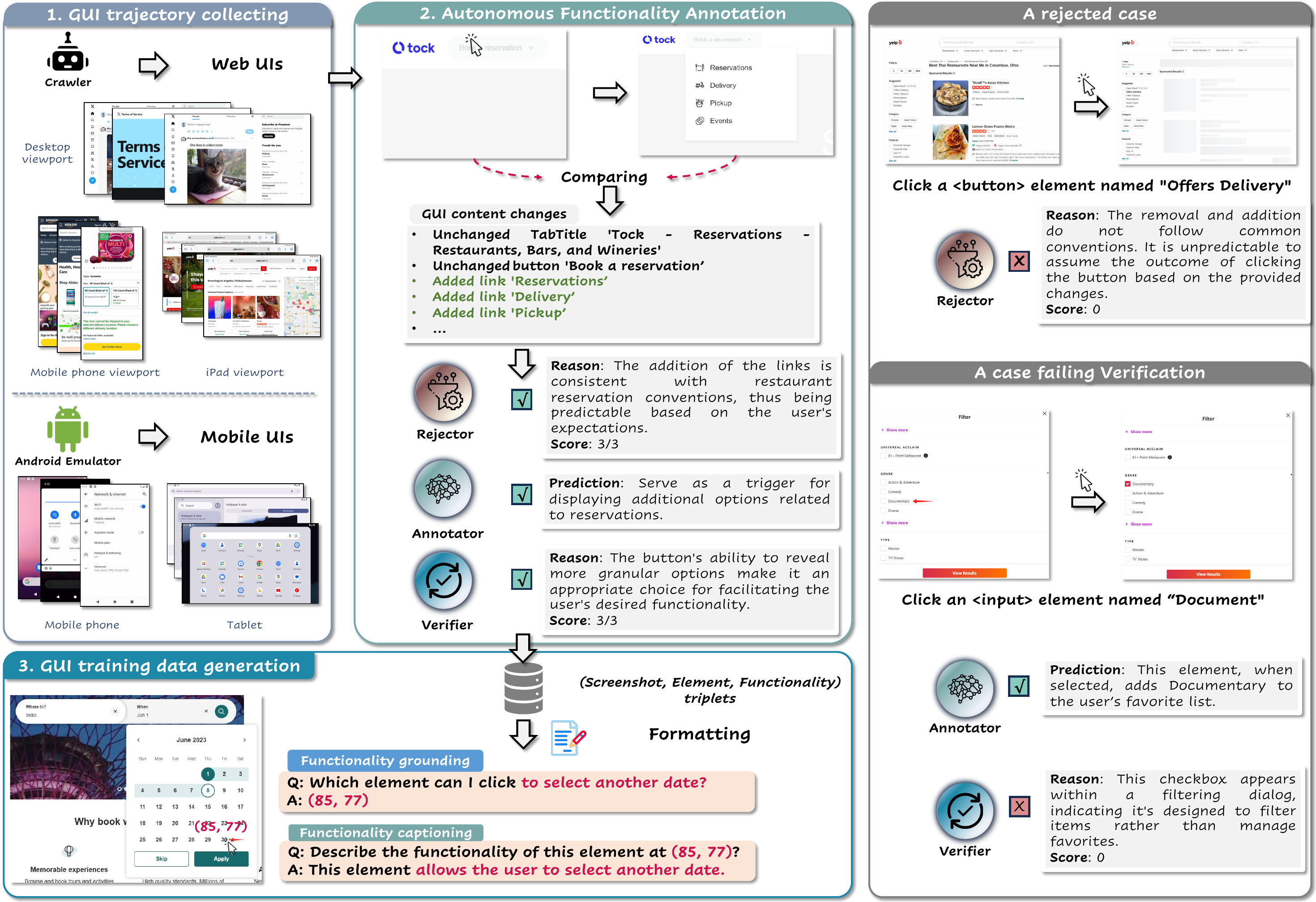}
    \caption{\textbf{The proposed pipeline for automatic UI functionality annotation.} An LLM is utilized to predict element functionality based on the UI content changes observed during the interaction. LLM-aided rejection and verification are introduced to improve data quality. Finally, the high-quality functionality annotations will be converted to instruction-following data by applying task templates.}
    \label{fig: anno pipeline}
\end{figure*}

This section introduces \methodname{}, an annotation pipeline (Fig.~\ref{fig: anno pipeline}) that automatically produces contextual element functionality annotations used to enhance VLMs' GUI grounding capabilities.

\subsection{Collecting UI Interaction Trajectories}
Our pipeline initiates by collecting interaction trajectories, which are sequences of UI contents captured by interacting with UI elements. Each step captures all interactable elements and the accessibility tree (AXTree) that briefly outlines the UI structure, which will be used to annotate functionality. To amass these trajectories, we utilize the latest Common Crawl repository as the data source for web UIs and Android Emulator for mobile UIs. The open-source trajectories from AndroidControl~\citep{androidcontrol} and MobileViews~\citep{mobileviews} are also included to enhance diversity. Note that illegal UIs are manually excluded from the sources. Please refer to Sec.~\ref{sec:supp:data details} for collecting details and data license.

\subsection{Automatic Functionality Annotation}
The pipeline generates functionality annotations for elements in the collected trajectories. Interacting with an element $e$, by clicking or hovering over it, triggers UI content changes. In turn, these changes can be used to predict the functionality $f$ of the interacted element. For instance, if clicking an element causes new buttons to appear in a column, the element likely functions as a dropdown menu activator (an example in Fig.~\ref{fig: funcpred diff case}).
With this observation, we utilize a capable LLM (i.e., Llama-3-70B~\citep{llama3modelcard}) as a surrogate for humans to summarize an element's functionality based on the UI changes resulting from interaction. Concretely, we generate compact content differences for AXTrees before ($s_t$) and after ($s_{t+1}$) the interaction using a file-comparing library\footnote{https://docs.python.org/3/library/difflib.html}. Then, we prompt the LLM to analyze the UI content changes (addition, deletion, and unchanged lines), present a detailed Chain-of-Thoughts~\citep{wei2022chain} reasoning process explaining how the element affects the UI, and finally summarize the element's functionality.

In cases where element interactions significantly transform the UI and cause lengthy differences—such as navigating to a new screen—we adjust our approach by using UI description changes instead of the AXTree differences.

\begin{figure*}[t]
    \centering
    \includegraphics[width=1.0\linewidth]{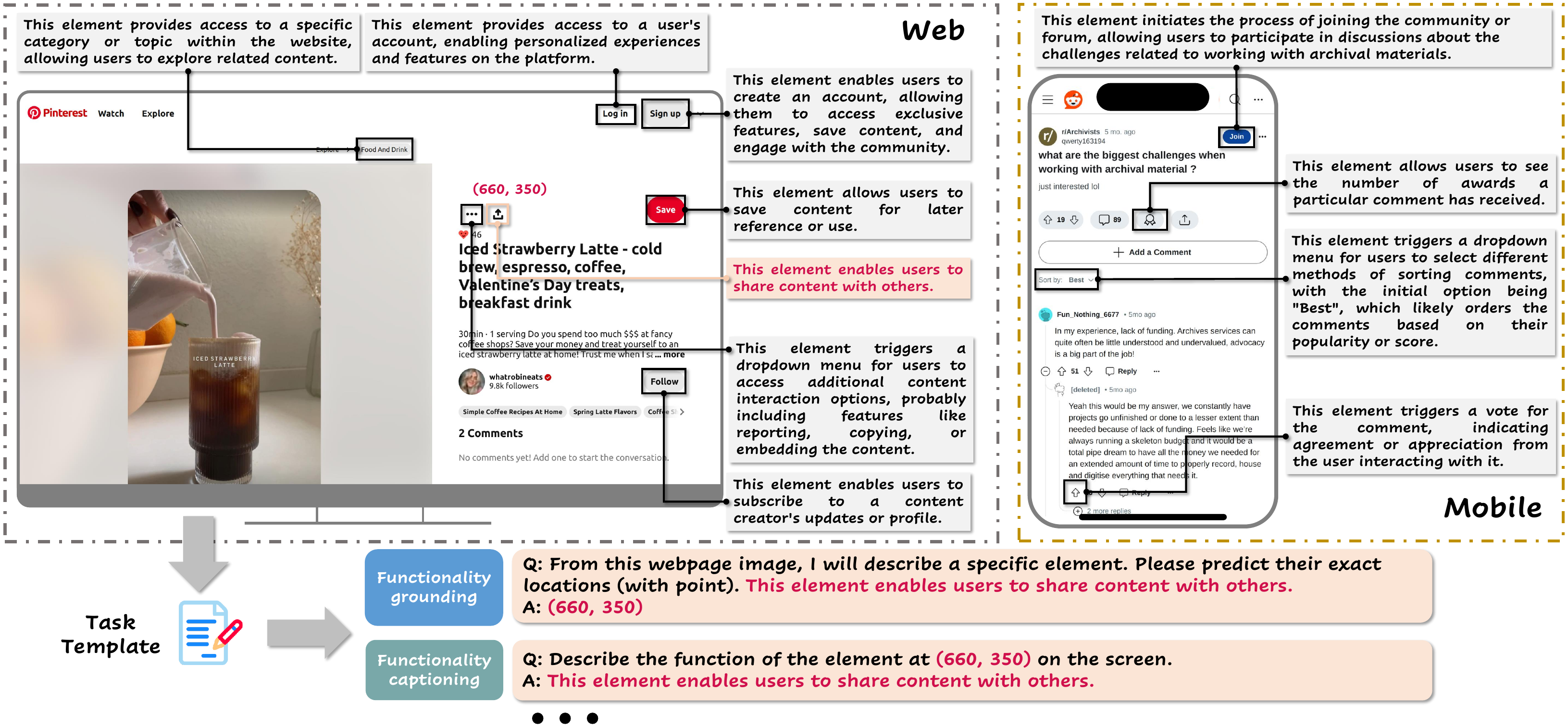}
    \caption{Element functionality annotations generated by the AutoGUI pipeline for both web and mobile domains.}
    \label{fig: our dataset}
\end{figure*}

This annotation process is formulated as: $f = \text{LLM}(p_{\text{anno}}, s_t, s_{t+1})$ where $f$ is the predicted functionality, $p_{\text{anno}}$ is the annotation prompt (Tab.~\ref{tab:supp:funcpred manip prompt} and Tab.~\ref{tab:supp:funcpred nav prompt}). Examples are depicted in Fig.~\ref{fig: our dataset} and more annotation details are explained in Sec.~\ref{sec:supp:anno details}.

\subsection{Removing Invalid Samples via LLM-Aided Rejection}
The collected trajectories may contain invalid samples due to broken UIs, such as incomplete UI loading, which can mislead the models trained with them. To filter out these invalid samples, we introduce an LLM-aided rejection approach. Initially, hand-written rules (detailed in Sec.~\ref{sec:supp:hand-written rules}) are used to detect obvious bad cases, such as blank UIs, UIs containing elements indicating content loading, and interaction targets outside of UIs. However, a few types are difficult to detect with the rules. For instance, interacting with a “view more” button might unexpectedly redirect the user to a login page instead of the desired information page due to website login restrictions. To identify these challenging samples, we prompt the annotating LLM to also act as a rejector. Specifically, the LLM takes the UI content changes as input, provides detailed reasoning through whether the changes are meaningful for predicting the element's functionality, and finally outputs predictability scores ranging from 0 to 3 (3 is empirically chosen for a balance between annotation efficiency and quality.). This process is formulated as follows: $score = \text{LLM}(p_{\text{reject}}, e, s_t, s_{t+1}) $
where $p_{\text{reject}}$ is the rejection prompt (Tab.~\ref{tab:supp:rejection prompt}).

This approach ensures that predictable samples receive higher scores, while unpredictable ones receive lower scores. For instance, if a button labeled "Show More", upon interaction, clearly adds new content, this sample will be considered to provide sufficient changes that can anticipate the content expansion functionality and will get a score of 3.

We deploy this rejector to discard the bottom 30\% of samples based on score ranking to strike a balance between the elimination of invalid samples and the preservation of valid ones (Details in Sec.~\ref{suc:supp:reject details}). The samples that pass the rejection procedure are submitted for functionality annotation. Please see examples in Fig.~\ref{fig: rejection examples}.

\subsection{Improving Annotation Quality via LLM-Based Verification}
The functionality annotations produced by the LLM probably contain incorrect and hallucinated samples. To improve dataset quality, we prompt LLMs to verify the annotations, inspired by works that justify the feasibility of LLM-based verification~\citep{zheng2023judging,Chen_Qin_Jiang_Choi_2024,lee-etal-2023-making}. This process presents the LLMs with the interacted element, its UI context, the UI changes induced by the interaction, and the functionality annotation generated in the previous stage. Then, the LLMs analyze the UI content changes and predict whether the interacted element aligns with the given functionality. If the LLMs determine that the interacted element fulfills the functionality given its UI context, the LLMs will grant a full score (An example in Fig.~\ref{fig: verif diff case}). If not, this functionality will be seen as incorrect as this mismatch indicates that it may not accurately reflect the element's role within the UI context.

To mitigate the potential biases in LLMs~\citep{panickssery2024llm, bai2024benchmarking}, two different LLMs (i.e., Llama-3-70B and Mistral-7B-Instruct-v0.2) are employed as verifiers and prompted to output 0-3 scores (This scoring range is chosen as it empirically achieves a high verification accuracy). The scoring process is formulated as follows: $score = \text{LLM}(p_{\text{verify}}, e, f, s_t, s_{t+1}) $
where $p_{\text{verify}}$ denotes the verification prompt (Tab.~\ref{tab:supp:verif prompt}). Only if the two scores are both 3s do we consider the functionality annotation correct (Details in Sec.~\ref{suc:supp:verif details}). While this approach seems stringent, we can make up the number of annotations through scaling.

\begin{table*}[]
\centering
\small
\caption{\textbf{The statistics of the AutoGUI datasets.} The Anno. Tokens and Avg. Words columns show the total number of tokens and the average number of words for the functionality annotations regardless of task templates. The Domains/Apps column shows the number of unique web domains/mobile Apps involved in each split.}
\label{tab:simple data stats}
\begin{tabular}{@{}ccccccc@{}}
\toprule
Split & \#Tasks & Anno. Tokens & Avg. Words & Domains/Apps & Device Ratio   \\                                                                   \midrule
Train & 702k  & 17.9M        & 23.1       & 916     & Web: $54.6\%$, Mobile: $45.4\%$                                              \\ \cmidrule(r){1-6}
Test  & 2k    & 53.4k        & 22.5       & 299     & Web: $50\%$, Mobile: $50\%$                                                                                                               \\ \bottomrule
\end{tabular}
\end{table*}

\begin{figure*}[t]
    \centering
    \includegraphics[width=1.0\linewidth]{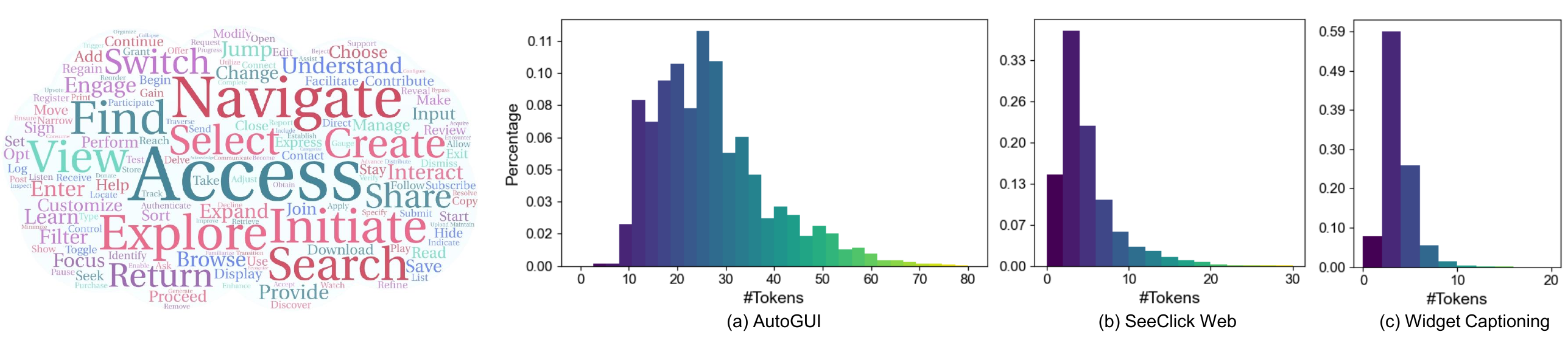}
    \caption{\textbf{Diversity of the AutoGUI dataset.} \textbf{Left}: The word cloud illustrates the ratios of the verbs representing the main intents in the functionality annotations. \textbf{Right}: Comparing the distributions of the annotation token numbers for our AutoGUI training split, SeeClick Web training data~\citep{cheng2024seeclick}, and Widget Captioning~\citep{Li2020WidgetCG}. The comparison demonstrates that our dataset covers significantly more diverse task lengths.}
    \label{fig: wordcloud and tokdistrib}
\end{figure*}

\subsection{Task Generation}
After rejecting, annotating, and verifying, we obtain a high-quality UI functionality dataset containing triplets of \{\textit{UI screenshot, Interacted element, Functionality}\}. To convert this dataset into an instruction-following dataset for training and evaluation, we generate functionality grounding and referring tasks using diverse prompt templates. The coordinates of element bounding boxes are normalized within the range $[0,999]$ (see Fig.~\ref{fig: our dataset}).

We finally annotate 2k grounding samples as a test set and 702k as a training set (The details of ensuring no overlap between the two sets can be seen in Sec.~\ref{sec:supp:data stats}). The statistics of our dataset in Tab.~\ref{tab:simple data stats} and Sec.~\ref{sec:supp:data stats} show that our dataset covers diverse UIs and exhibits variety in lengths and functional semantics of the annotations.

\subsection{Analysis of Data Quality}

\noindent{\textbf{Comparison with Human Annotation}} $N=145$ samples (99 valid and 46 invalid) are randomly selected as a testbed for comparing the annotation correctness of a trained human annotator and our pipeline. Here, correctness is defined as $Correctness = C / (N - R)$, where $C$ and $R$ denote the numbers of correctly annotated and rejected samples, respectively. The denominator subtracts the number of rejected samples as we are more interested in the percentage of correct samples after rejecting invalid samples. The authors rigorously evaluate the annotation results based on three criteria outlined in Fig.~\ref{fig: check criteria}.

Details can be found in Sec.~\ref{sec:supp:humaneval details}.

After experimenting with three runs, Tab.~\ref{tab:ablate autogui} shows that the AutoGUI pipeline achieves high correctness comparable to the trained human annotator (r6 vs. r1). Without rejection and verification (r2), AutoGUI is inferior as it cannot recognize invalid samples. Notably, simply using the rules written by the authors can improve the correctness, which is further enhanced with the LLM-aided rejector (r4 vs. r3). Moreover, utilizing the annotating LLM itself to self-verify its annotations helps AutoGUI surpass the trained annotator (r5 vs. r1). Introducing another LLM verifier (i.e., Mistral-7B-Instruct-v0.2) brings a slight increase which results from Mistral recognizing Llama-3-70B’s incorrect descriptions of how dropdown menu options work. Overall, these results justify the efficacy of the AutoGUI annotation pipeline.

Qualitative comparison (Fig.~\ref{fig: autogui vs human}) shows that our pipeline generates more detailed annotations which would take more time for the human annotator.

\noindent{\textbf{Impact of LLM Output Uncertainty}} Despite LLM output uncertainty, our pipeline achieves a high annotation consistency of 94.5\%. LLM uncertainty affects rejection but has a minimal overall impact due to the low prevalence of invalid samples. More experimental details in the Appendix.

\begin{table}[]

\centering
\caption{\textbf{Comparing AutoGUI and human annotator.} AutoGUI with the proposed rejection and verification achieves correctness comparable to the trained human annotator. One LLM means Llama-3-70B and Two LLMs include Mistral-7B-Instruct-v0.2.}
\label{tab:ablate autogui}
\resizebox{\columnwidth}{!}{%
\begin{tabular}{@{}ccccc@{}}
\toprule
No. & Annotator  & Rejector   & Verifier              & Correctness \\ \midrule
r1 & Human      & -          & -                     & 95.5\%      \\
r2 & Llama-3-70B & -          & -                     & 64.5\%      \\
r3 & Llama-3-70B & Rules      & -                     & 83.1\%      \\
r4 & Llama-3-70B & Rules+LLM  & -                     & 94.4\%      \\
r5 & Llama-3-70B & Rules+LLM  & One LLM            & 96.0\%      \\
r6 & Llama-3-70B & Rules+LLM & Two LLMs & \textbf{96.7\%}      \\ \bottomrule
\end{tabular}
}
\end{table}

\section{Fine-Tuning Experiments}
\begin{table*}[]
\centering
\scriptsize

\caption{\textbf{Element grounding accuracy across benchmarks.} We compare the base models fine-tuned with our AutoGUI data and representative open-source VLMs. General-purpose models (Qwen-VL, SliME-8B, and Qwen2-VL) show significant performance improvements after fine-tuning with the AutoGUI data. The UI-specialized model (SeeClick) also improves when AutoGUI data is added to their fine-tuning datasets. Green text indicates gains over the base models. $\dag$ denotes metrics quoted from the original benchmark paper.}
\label{tab:eval results}
\resizebox{2\columnwidth}{!}{%
\begin{tabular}{ccc|cccccc}
\hline
Type                        & Model                                                    & Size & FuncPred      & ScreenSpot                                 & ScreenSpot-v2 & MoTIF         & VWB EG                        & VWB AG        \\ \hline
\multirow{5}{*}{General}    & GPT-4o                                                   & N/A  & 9.8           & 17.8                                       & 20.4          & 30.5          & 5.6                           & 6.8           \\
                            & Llama-3.2-Vision-Instruct                                & 11B  & 4.9           & 11.7                                       & 11.6          & 19.7          & 7.0                           & 3.9           \\
                            & SliME~\citep{slime}                & 8B   & 3.2           & 13.0                                       & 13.4          & 7.0           & 6.1                           & 4.9           \\
                            & Qwen-VL~\citep{bai2023qwen}        & 10B  & 3.0           & 5.2$^{\dag}$                               & 5.6           & 7.8           & 1.7                           & 3.9           \\
                            & Qwen2-VL~\citep{bai2023qwen}       & 7B   & 38.7          & 66.4                                       & 66.9          & 71.1          & 55.9                          & 62.1          \\ \hline
\multirow{4}{*}{UI Experts} & CogAgent~\citep{hong2023cogagent}  & 18B  & 29.3          & 47.4$^{\dag}$                              & 52.1            & 45.1 & 55.7 & 59.2 \\
                            & SeeClick~\citep{cheng2024seeclick} & 10B  & 19.8          & 53.4$^{\dag}$ & 54.0 & 66.5          & 39.2                          & 27.2          \\
                            & UGround-v1-7B~\citep{uground}      & 7B   & 55.8          & 85.9                                       & 88.0 & 78.4          & 92.7                          & 69.9          \\
                            & OS-ATLAS~\citep{osatlas}           & 7B   & 52.1          & 82.5                                       & 84.1 & 78.8          & 82.6                          & 71.8          \\ \hline
\multirow{5}{*}{Finetuned}  & Qwen-VL-AutoGUI702k                                      & 10B  & 48.7 \gain{(+45.7)} & 41.2 \gain{(+36.0)}                                    & 40.2 \gain{(+34.6)}           & 44.0 \gain{(+36.2)}        & 42.1 \gain{(+40.4)}                        & 35.9 \gain{(+32.0)}         \\ 
                            & SliME-AutoGUI702k                                        & 8B   & 62.6 \gain{(+59.4)} & 44.0 \gain{(+31.0)}                     & 42.5 \gain{(+29.1)}         & 44.9 \gain{(+37.9)}         & 25.4 \gain{(+19.3)}                          & 13.6 \gain{(+8.7)}         \\
                            & Qwen2-VL-AutoGUI702k                                  & 7B   & 65.0 \gain{(+26.3)}& 80.0 \gain{(+13.6)}                               & 83.2 \gain{(+16.3)}   & 72.3 \gain{(+1.2)}    & 90.3 \gain{(+34.4)}& 70.9 \gain{(+8.8)}   \\
                            & SeeClick w/ AutoGUI702k                                  & 10B   & 50.0 \gain{(+30.2)}  & 54.2  \gain{(+0.8)}                             & 54.7 \gain{(+0.7)}     & 67.0 \gain{(+0.5)}          & 56.2 \gain{(+17.0)}                 & 45.6 \gain{(+18.4)}         \\ 
                            \hline
\end{tabular}%
}
\end{table*}
\begin{figure}[h]
    \centering
    \includegraphics[width=1.0\linewidth]{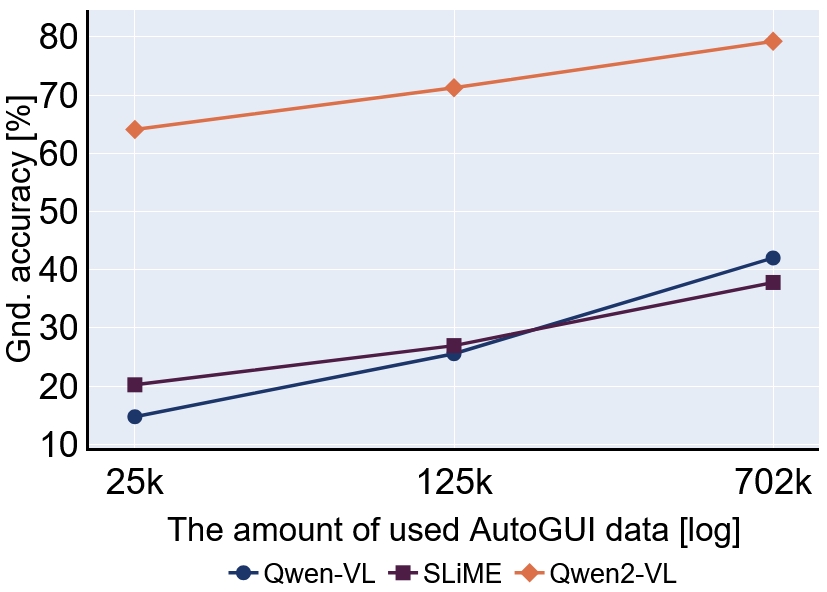}
    \caption{\textbf{Scaling effect of the AutoGUI data.} The three general-purpose VLMs are fine-tuned with three scales of AutoGUI data. Using more data consistently enhances the grounding accuracy of the three models. Note that the grounding accuracy (Y-axis) is averaged over all the element grounding benchmarks.}
    \label{fig: gnd scaling}
\end{figure}

\vspace{-2mm}
This section validates that our dataset effectively enhances the GUI grounding capabilities of VLMs.
\vspace{-1mm}
\subsection{Experimental Settings}
\vspace{-1mm}
\noindent\textbf{Evaluation Benchmarks} We base our evaluation on the UI grounding benchmarks for various scenarios: \textbf{FuncPred} is the test split from our collected functionality dataset. This benchmark requires a model to locate the element specified by its functionality description. \textbf{ScreenSpot}~\citep{cheng2024seeclick} and \textbf{ScreenSpot-v2}~\citep{osatlas} require a model to locate elements based on short instructions on mobile, desktop, and web platforms.

\textbf{VisualWebBench (VWB)}~\citep{liu2024visualwebbench} is a comprehensive multi-modal benchmark assessing the understanding capabilities of VLMs in web scenarios. We select the element and action grounding tasks from this benchmark.

\textbf{MOTIF}~\citep{Burns2022ADF} requires an agent to complete a natural language command in mobile Apps. Samples of the benchmarks are visualized in Fig.~\ref{fig: bmk samples}.
We report the grounding accuracy (\%): $\text { Acc }= \sum_{i=1}^N \mathbf{1}\left(\text {pred}_i \text { inside GT } \text {bbox}_i\right) / N \times 100 $ where $\mathbf{1}$ is an indicator function and $N$ is the number of test samples. This formula denotes the percentage of samples with the predicted points lying within the bounding boxes of the target elements.

\noindent\textbf{Training Details}
We select SliME-8B~\citep{slime}, Qwen-VL~\citep{bai2023qwen}, and Qwen2-VL-7B~\citep{qwen2vl} as the base models and fine-tune them on 25k, 125k, and 702k samples of the AutoGUI training data to investigate how the AutoGUI data enhances their UI grounding capabilities. We fine-tune Qwen-VL and Qwen2-VL with LoRA~\citep{hu2022lora} and fine-tune SliME~\citep{slime} with only the visual encoder frozen. We also test the benefits of our dataset for a UI expert VLM, i.e., SeeClick~\citep{cheng2024seeclick}, by adding our data to its fine-tuning data. All models are fine-tuned on 8 A100 GPUs for one epoch. (More details and hyper-parameters in Sec.~\ref{sec:supp:impl details})

\noindent\textbf{Compared VLMs}
We compare with both general-purpose VLMs and UI expert VLMs. During the evaluation, we manually craft grounding prompts suitable for these VLMs.

\subsection{Experimental Results and Analysis}

\begin{table}[!t]
\centering
\footnotesize
\caption{\textbf{Comparing the AutoGUI functionality annotation type with existing types}. Qwen-VL is separately fine-tuned with the four annotation types. Our functionality annotation leads to superior grounding accuracy.}
\label{tab:ablation}
\resizebox{\columnwidth}{!}{%
\begin{tabular}{@{}ccccc@{}}
\toprule
Data Size             & Variant          & FuncPred & MOTIF & ScreenSpot \\ \midrule
\multirow{4}{*}{25k}  & w/ Elem-HTML     &  5.3      &  11.7   &    5.7\\
                      & w/ Condensed Func.     &  3.8&  19.8  &   4.8\\
                      & w/ Func. (Ours full)         &    \textbf{21.1}    &   \textbf{22.5}   &   \textbf{16.4}    \\ \midrule
\multirow{4}{*}{125k} & w/ Elem-HTML     &  15.5   &  15.8  &   17.0      \\
                      & w/ Condensed Func.     &  14.1   &  23.7  &   23.8      \\
                      & w/ Func. (Ours full)         &  \textbf{24.6}   &  \textbf{28.7}  &   \textbf{27.0}    \\ \bottomrule
\end{tabular}
}
\end{table}

\begin{figure*}[th]
    \centering
    \includegraphics[width=\linewidth]{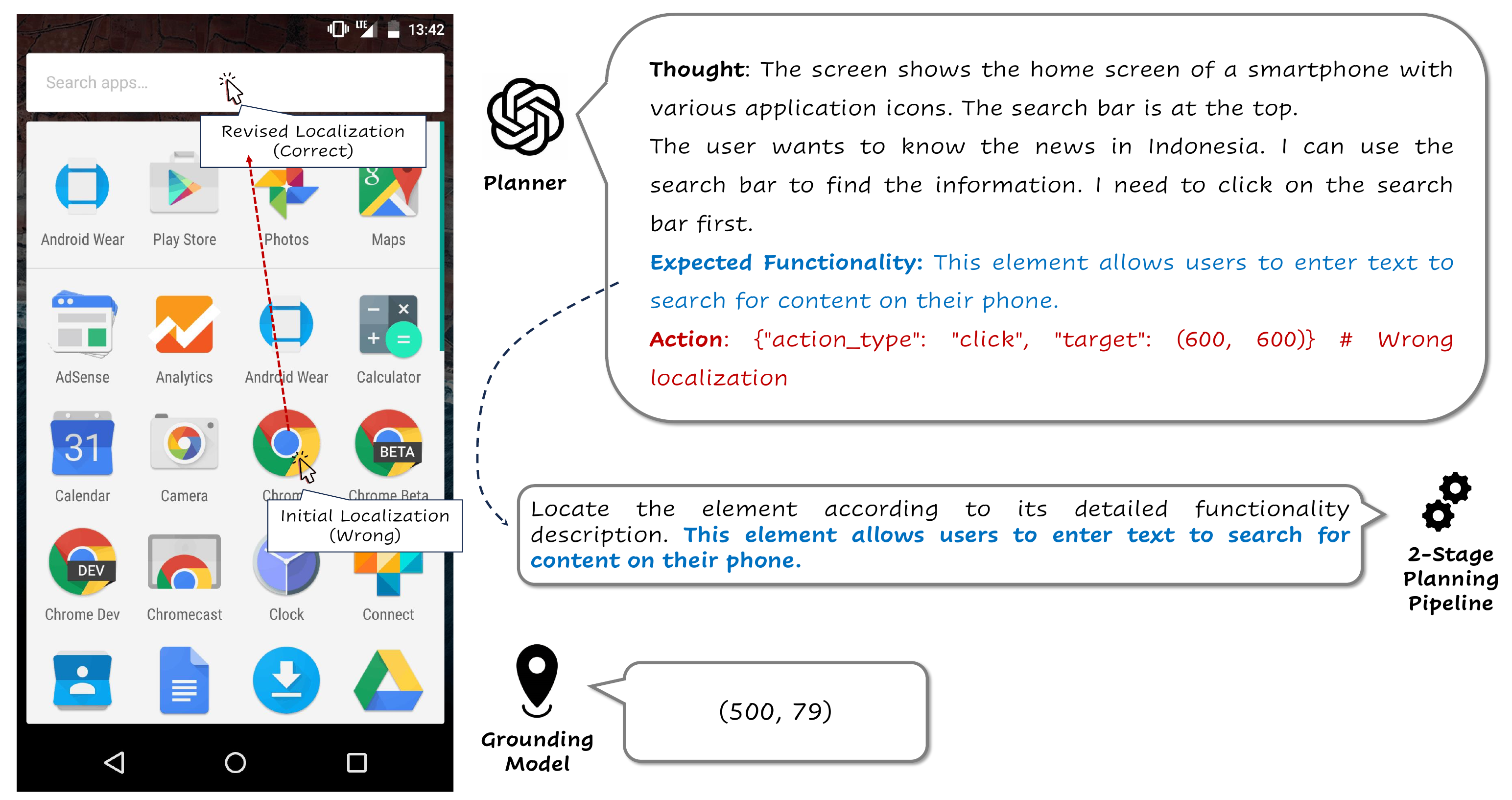}
    \caption{\textbf{An example of the 2-stage planning setting used in Sec.~\ref{sec:2stage planning} to demonstrate the potential use of AutoGUI data.} The planner (a proprietary VLM, e.g., Gemini) is bad at outputting numeric coordinates when locating elements. The grounding model, finetuned with AutoGUI functionality grounding tasks, can correctly locate the task-related target element according to the expected functionality description output by the planner.}
    \label{fig: 2stage example}
\end{figure*}

\begin{table*}[th]
\centering
\caption{\textbf{Applying our AutoGUI dataset to 2-stage GUI agent task planning on AITW benchmark.} The results show that Qwen2-VL trained with AutoGUI functionality grounding tasks can overtake the element grounding process of the proprietary models to achieve significantly higher step accuracy. \textbf{Step Acc.} means the percentage of correctly planned actions while \textbf{Click acc.} means the percentage of correctly planned click actions.}
\label{tab:2-stage planning}
\resizebox{2\columnwidth}{!}{%
\begin{tabular}{@{}c|c|cccccc@{}}
\toprule
Planner                               & Grounding Model            & \textbf{\begin{tabular}[c]{@{}c@{}}General\\ Step acc. / Click acc.\end{tabular}} & \textbf{\begin{tabular}[c]{@{}c@{}}Install\\ Step acc. / Click acc.\end{tabular}} & \textbf{\begin{tabular}[c]{@{}c@{}}Google Apps\\ Step acc. / Click acc.\end{tabular}} & \textbf{\begin{tabular}[c]{@{}c@{}}Single\\ Step acc. / Click acc.\end{tabular}} & \textbf{\begin{tabular}[c]{@{}c@{}}Webshopping\\ Step acc. / Click acc.\end{tabular}} & \textbf{Avg Step acc.} \\ \midrule
\multirow{2}{*}{GPT-4o-mini}          & GPT-4o-mini    & 14.85 / 9.58       & 11.17 / 5.76   & 12.08 / 6.85      & 21.09 / 11.24   & 10.89 / 11.22  & 14.01       \\
                                      & Qwen2-VL-7B SFT w/ AutoGUI & 20.43 / 20.56 & 25.59 / 22.49 & 15.25 / 12.33 & 25.59 / 22.49    & 16.15 / 20.53   & \textbf{18.37} \gain{(+4.36)}     \\ \midrule
\multirow{2}{*}{Gemini-2.0-flash-exp} & Gemini-2.0-flash-exp       & 26.37 / 18.16                                                                    & 28.49 / 26.91                                                                    & 30.30 / 22.88                                                                        & 41.94 / 28.95                                                                   & 20.22 / 22.65                                                                        & 29.50       \\
                                      & Qwen2-VL-7B SFT w/ AutoGUI & 36.34 / 36.54                                                                    & 50.95 / 48.95                                                                    & 40.99 / 40.52                                                                        & 50.95 / 48.95                                                                   & 32.83 / 43.52                                                                        & \textbf{39.23} \gain{(+9.73)}     \\ \bottomrule
\end{tabular}%
}
\end{table*}

\noindent\textbf{A) AutoGUI functionality annotations effectively enhance VLMs' UI grounding capabilities and achieve scaling effects.} We endeavor to show that the element functionality data collected by AutoGUI contributes to high grounding accuracy. The results in Tab.~\ref{tab:eval results} demonstrate that the base models embrace notable performance gains on all the benchmarks.  The two general-purpose VLMs (Qwen-VL and SLiME), which perform poorly, witness huge performance increases after fine-tuning with AutoGUI data. Qwen2-VL fine-tuned with our data achieves high accuracy comparable to the expert models, i.e., UGround and OS-ATLAS. Interestingly, the UI expert VLM (i.e., SeeClick) also benefits from our data, with remarkable performance gains on FuncPred and VWB.

Fig.~\ref{fig: gnd scaling} shows that the three general-purpose VLMs obtain progressively rising grounding accuracy as the AutoGUI data size scales from 25k to 702k, indicating that increasing AutoGUI data amount leads to better localization performance.

In summary, our functionality data enhances VLMs UI element grounding ability and exhibits clear scaling effects as the data size increases.

\noindent\textbf{B) Our functionality annotations are effective for enhancing UI grounding capabilities.} To assess the effectiveness of functionality annotations, we compare this annotation type with three types: 1) \textbf{Naive element-HTML pairs}, which are directly obtained from the UI source code~\citep{hong2023cogagent} and associate HTML code with elements in specified areas of a screenshot. Examples are shown in Fig.~\ref{fig: functionality vs others}. To create these pairs, we replace the functionality annotations with the corresponding HTML code snippets recorded during trajectory collection. 2) \textbf{Brief functionality descriptions} that are generated by prompting GPT-4o-mini\footnote{https://openai.com/index/gpt-4o-mini-advancing-cost-efficient-intelligence/} to condense the AutoGUI functionality annotations. For example, a full description such as \textit{`This element provides access to a documentation category, allowing users to explore relevant information and guides'} is shortened to \textit{`Documentation category access'}.

After experimenting with Qwen-VL~\citep{bai2023qwen} at the 25k and 125k scales, the results in Tab.~\ref{tab:ablation} show that fine-tuning with the complete functionality annotations is superior to the other three types. Notably, our functionality annotation type yields the largest gain on the challenging FuncPred benchmark that emphasizes contextual functionality grounding.
In contrast, the Elem-HTML type performs poorly due to the noise inherent in HTML code (e.g., numerous redundant tags), which reduces fine-tuning efficiency. The condensed functionality annotations are also inferior, as the consensing loses details necessary for fine-grained element grounding. In summary, the AutoGUI functionality annotations provide a clear advantage in enhancing UI grounding capabilities.

\subsection{Grounding Failure Case Analysis}

After analyzing the grounding failure cases, we identified several failure patterns in the fine-tuned models: a) difficulty in accurately locating small elements; b) challenges in distinguishing between similar but incorrect elements; and c) issues with recognizing icons that have uncommon shapes. Please refer to Sec.~\ref{sec:supp:case analysis} for details.

\section{Potential Use of AutoGUI Data}
\label{sec:2stage planning}

We apply our dataset to a downstream GUI agent task to demonstrate how our dataset can benefit GUI agents based on proprietary VLMs. The used benchmark is AITW~\citep{rawles2023android} which requires an agent to complete high-level user instructions on mobile apps. The metric is step accuracy, where a planned step is considered correct only if the action type and arguments match ground truths.

\noindent \textbf{2-stage planning} Following UGround~\citep{uground}, a planner model initially performs reasoning through task progress and UI content and then plans the next action. We prompt the planner to also describe the expected functionality of the target element for click actions. Next, a grounding model (Qwen2-VL-7B) trained with our functionality grounding tasks is used to locate the target according to the functionality description. See an example in Fig.~\ref{fig: 2stage example} and additional details in Sec.~\ref{sec:supp:2stage planning}).

The results in Tab.~\ref{tab:2-stage planning} show that even the strong proprietary VLMs (e.g. Gemini) possess weak UI element grounding capability. Qwen2-VL trained with AutoGUI functionality grounding tasks can overtake the element grounding process of the proprietary models and help the planners achieve significantly higher step accuracy by correcting the target locations of the \textit{click} actions.

Although this experiment is not designed to surpass expert models tailored for agent tasks, we hope it can facilitate further research of GUI agents with strong element grounding ability.

\section{Conclusion}
We propose \methodname{}, a scalable and automatic annotation pipeline aimed to produce massive UI element functionality annotations used to enhance UI grounding capabilities of VLMs. The pipeline prompts an open-source LLM to generate element functionalities based on the UI content changes induced by interacting with the elements. LLM-aided rejection and verification are used to guarantee high quality. Fine-tuned with the data collected by \methodname{}, the base models obtain stronger UI grounding ability and exhibit data scaling effects. We hope that \methodname{} will open up possibilities for advancing the field of UI agents.

\section*{Limitations}
AutoGUI is dedicated to providing an autonomous way to collect scalable UI grounding/captioning data for training capable UI-VLMs. However, AutoGUI still encounters several limitations:

\noindent\textbf{Lack of Diverse Mobile App Data.} As many Apps implement anti-emulator code, it is extremely difficult to navigate through popular Apps, such as TikTok and WeChat, on Android emulators. To circumvent this issue, AutoGUI renders webpages at various resolutions, including smartphone resolution, to mimic diverse device types. Although mainstream websites, such as YouTube and Reddit, provide delicately designed webpage responsiveness for various resolutions, a number of less common websites do not possess such flexible responsiveness and distort severely when rendered at smartphone resolutions. Therefore, collecting UI data at a smartphone resolution probably leads to domain gaps between the collected data and real smartphone Apps that are not rendered with HTML.

\noindent\textbf{AutoGUI is Not Indented to Record Task-Oriented Interaction Trajectories.} AutoGUI randomly interacts with UIs to record transition trajectories and utilize the UI content changes to predict the functionalities of the interacted elements. Hence, the collected trajectories do not provide high-level task semantics. In other words, the AutoGUI dataset does not contain tasks that combine multiple low-level steps, such as selecting a check-in date and then a check-out date. These long-horizon tasks are usually generated by human annotators in the existing works~\cite{deng2024mind2web,rawles2023android}. In future work, we can also utilize capable LLMs to generate high-level tasks and then prompt the LLMs to interact with UIs according to the tasks.

\noindent\textbf{AutoGUI Cannot Annotate UI Elements That Modify Content on The Internet.} To avoid causing potential contamination on the Internet and bearing unexpected responsibilities, we try our best to eliminate interaction samples that manipulate sensitive elements that probably modify contents on the Internet. For example, elements used to post comments, make purchases, and enter account information are discarded. Consequently, the AutoGUI pipeline mainly annotates elements that only support read-only functionalities.

\section*{Acknowledgments}
This work was supported in part by the National Key R\&D Program of China (No. 2022ZD0160102), the National Natural Science Foundation of China (No. U21B2042, No. 62320106010), and in part by the 2035 Innovation Program of CAS.


\bibliography{AutoGUI_ACL2025}

\appendix

\counterwithin{figure}{section}
\counterwithin{table}{section}
\renewcommand\thefigure{\Alph{figure}}
\renewcommand\thetable{\Alph{table}}
\renewcommand{\thesubsection}{\Alph{subsection}}
\newpage
\section{Appendix}

\startcontents

{
\hypersetup{linkcolor=black}
\printcontents{}{1}{}
}
\clearpage

The appendix comprises the following sections:

Section A: Details for implementation details for the autonomous annotation pipeline, including dataset statistics, visualized annotation pipeline, and LLM prompts.

Section B: Details for model implementation and training.

Section C: Details of potential use of the AutoGUI dataset.

Section D: Additional experimental analysis including analysis of successful and failure cases on two benchmarks.

Section E: Limitations and Potential Societal Impact.

\subsection{Details of the AutoGUI Pipeline}
\label{sec:supp:data details}

\subsubsection{Extra Statistics of the AutoGUI Dataset}
\label{sec:supp:data stats}

\begin{figure*}[t]
    \centering
    \includegraphics[width=0.8\linewidth]{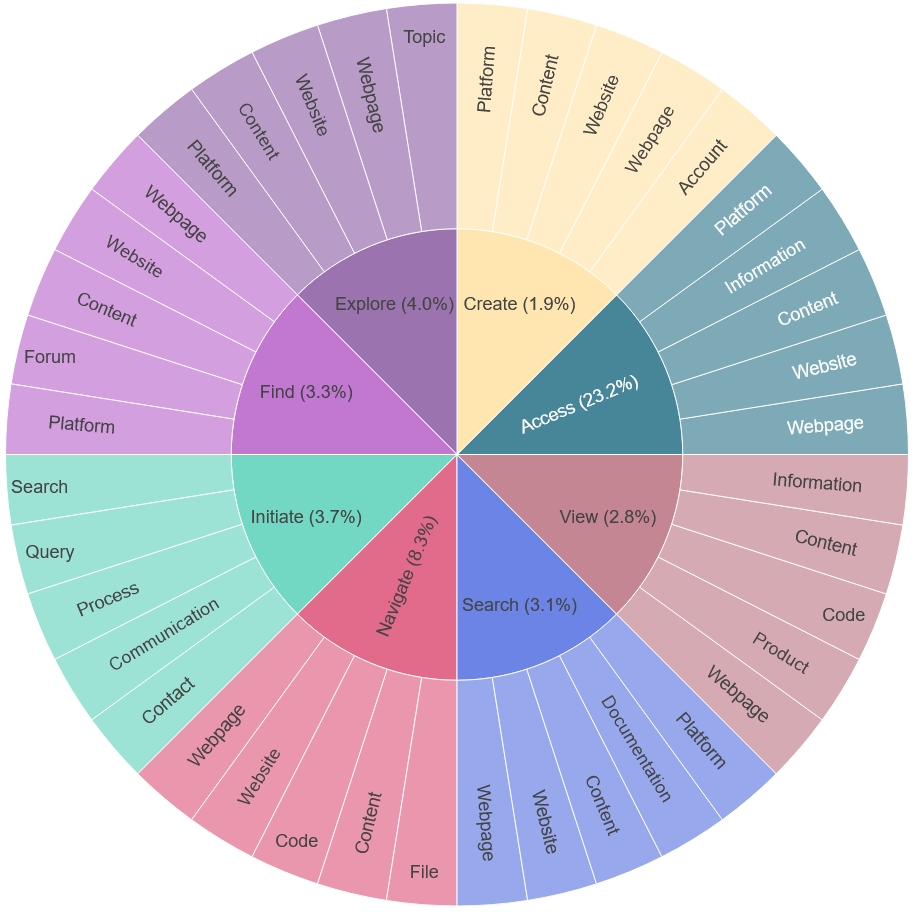}
    \caption{\textbf{Diversity of the verb-noun phrases of the AutoGUI dataset.} The top 10 verbs and their top 5 following nouns are displayed. This diagram shows that our dataset contains diverse tasks that involve various UI functions.}
    \label{fig: verbnoun}
\end{figure*}
\begin{figure*}[h]
    \centering
    \includegraphics[width=1.0\linewidth]{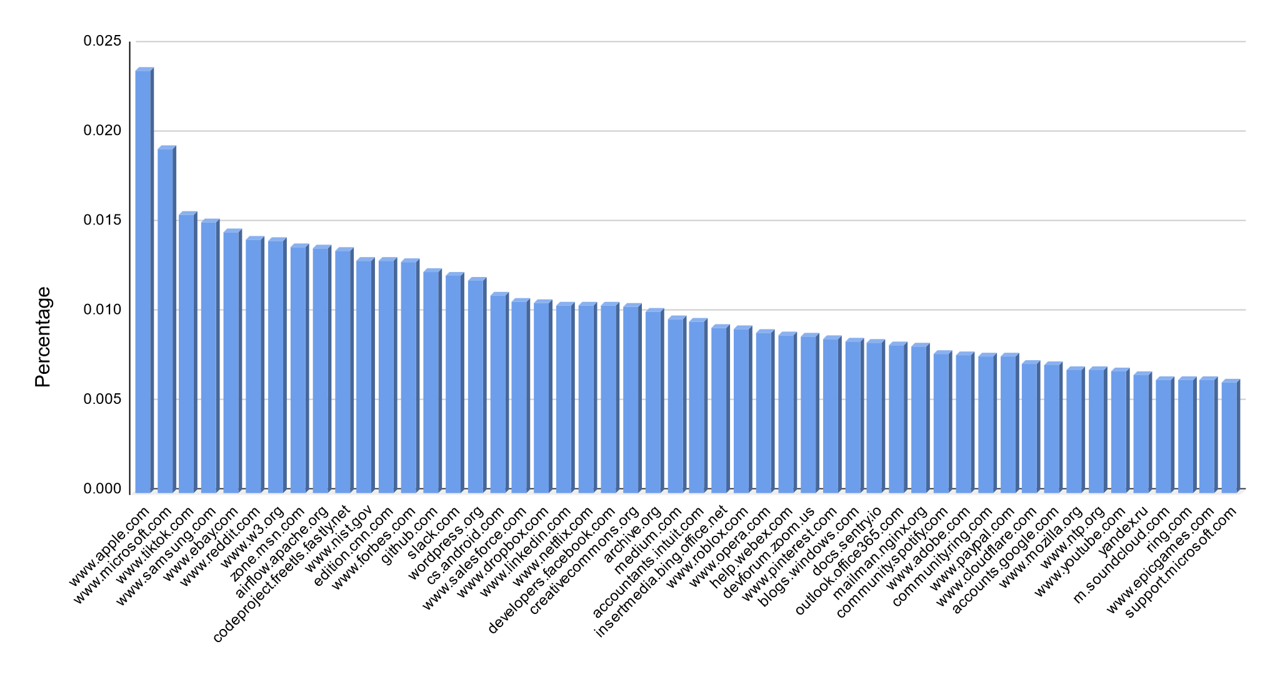}
    \caption{The top-50 most frequent top-level domains in the AutoGUI dataset.}
    \label{fig: domain ratio}
\end{figure*}

Fig.~\ref{fig: verbnoun} visualizes the verb-noun statistics of the AutoGUI dataset, highlighting its extensive coverage of diverse UI functionalities. Fig.~\ref{fig: domain ratio} lists the top 50 most frequent top-level domains in the AutoGUI dataset, showing that the AutoGUI dataset involves a broad spectrum of real-world scenarios, including technology (e.g., apple.com), entertainment (e.g., tiktok.com), office (e.g., outlook.com), news (e.g., medium.org), and finance (e.g., paypal.com).

\textbf{The approach to avoidding overlap between train and test data:} Since our focus is on annotating contextual functionality for GUI elements, we define two elements as distinct if they serve different functions within their respective contexts. For example, two "magnifier" buttons on the same GUI might have different roles—one for zooming in and the other for searching. To ensure no contamination, we investigated whether the test elements appeared in the training set by checking if bounding box overlapping occurred on the same GUIs. After this analysis, we found no such overlap.

\subsubsection{License of The Used Artifacts}
The licenses of the data sources on which the AutoGUI dataset is built are listed in Tab.~\ref{tab:supp:license}. These sources are all allowed to be used for academic research.

\begin{table*}[]
\centering
\caption{License or terms for use and/or distribution of the used artifacts in this work.}
\label{tab:supp:license}
\resizebox{2\columnwidth}{!}{%
\begin{tabular}{@{}ccccc@{}}
\toprule
Artifacts                   & License    & URLs containing Term-of-Use or other license information                                                          &  &  \\ \midrule
Common Crawl                & CC BY      & https://commoncrawl.org/terms-of-use                                                                              &  &  \\
AndroidControl Trajectories & Apache 2.0 & https://github.com/google-research/google-research/tree/41e4f1cbe1db648feb518a60501f638d9c8b25f2/android\_control &  &  \\
Mobile Views Trajectories   & MIT        & https://huggingface.co/datasets/mllmTeam/MobileViews                                                              &  &  \\ \bottomrule
\end{tabular}%
}
\end{table*}

\subsubsection{Recording Interaction Trajectories on Web}
\label{sec:supp:record traj detail}
\begin{figure*}[h]
    \centering
    \includegraphics[width=1.0\linewidth]{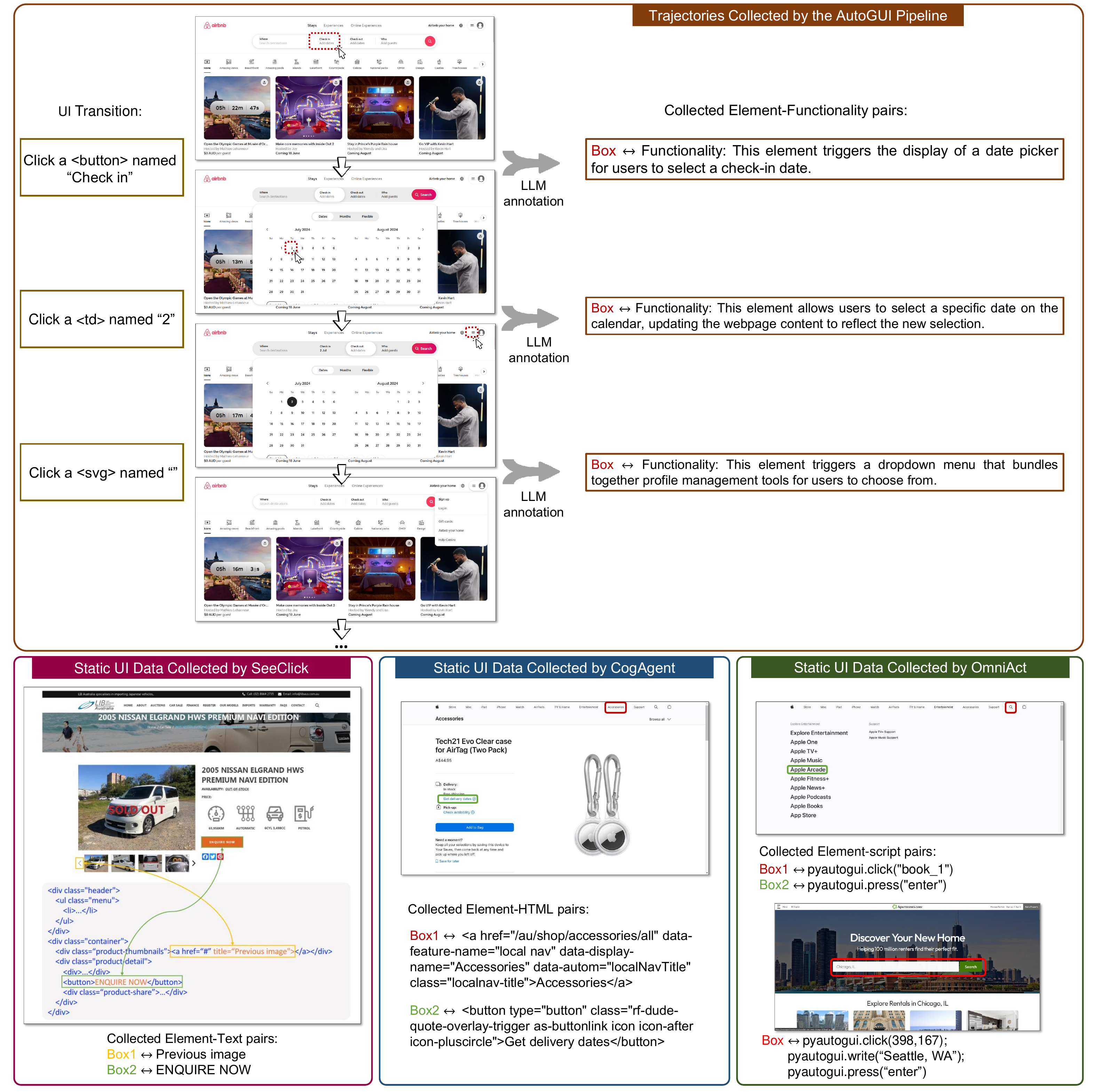}
    \caption{\textbf{Comparing the proposed AutoGUI annotation pipeline with existing methods.} AutoGUI is able to manipulate real UIs and interact with elements hidden beneath deeper levels (e.g., the buttons hidden in collapsed dropdown menus), thereby collecting considerably rich element-functionality annotations from the immense UI resources on the Internet. In contrast, SeeClick~\cite{cheng2024seeclick} only uses static webpages and collects static element-text pairs. Likewise, CogAgent collects static element-HTML pairs while OmniAct generates Python scripts only for visible elements. These three existing methods can only annotate visible static UI elements and ignore the rich UI functional semantics entailed in interaction trajectories which are provided by our AutoGUI pipeline in abundance.}
    \label{fig: pipeline comparison}
\end{figure*}

\noindent\textbf{Interactive Crawler for Common Crawl} We design an in-house web crawler that interacts with most elements rendered on the web page.
In contrast with existing methods which contain information for elements on the initial static web page for a given URL, our crawler randomly interacts with a rendered web page for \textbf{multiple steps} within a given action horizon $T_\text{act}$ to collect UI data with abundant functional semantics. Fig.~\ref{fig: pipeline comparison} compares the proposed AutoGUI and the existing annotation methods.
We empirically set $T_\text{act} = 10$ in all our recordings.
Therefore, our interactive crawler could collect functionality of elements that are not visible to static pages, including nested drop-down menus, date and location selectors, and secondary menus.

\noindent\textbf{Data Source and Data Format}
To incorporate a wide basis of web pages, we first obtain a list of the top-200 most visited domains~\footnote{\url{https://tranco-list.eu/}} and manually remove content delivery network (CDN) and not safe for work (NSFW) sites.
We use URLs in this curated list as seeds to query the Common Crawl index~\footnote{\url{https://index.commoncrawl.org/}} to find additional URLs with maximum sub-domain and path diversity.
Querying URLs from the Common Crawl index ensures that our crawler respects each site's robots.txt file, making the dataset collection process legally safe. 
By obeying the directives in robots.txt, we avoid potential legal issues associated with unauthorized web scraping.
For each web page, we collect the following data:
\begin{itemize}
    \item Screenshot image of the rendered page
    \item Accessible Tree (AXTree) text representing the page's accessibility structure
    \item HTML source code of the page
    \item Accessible Node (AXNode) text for the specific element our crawler interacted with at each step
\end{itemize}

\subsubsection{Recording Interaction Trajectories on Android Devices}
We also implement an in-house crawler that interacts with multiple emulated Google Pixel phones. The phones are reset to different starting UIs before a script randomly interacts with these phones to record trajectories. To improve data diversity, the starting UIs include the home page, drop-down panel, settings page, and Apps drawer.

Similar to webpage HTML, mobile phone UIs are rendered with XML code, which is cleaned and converted to AXTree-like content before being used to annotate functionalities.

\subsubsection{Functionality Annotation Details}
\label{sec:supp:anno details}
\begin{figure*}[th]
    \centering
    \includegraphics[width=1.0\linewidth]{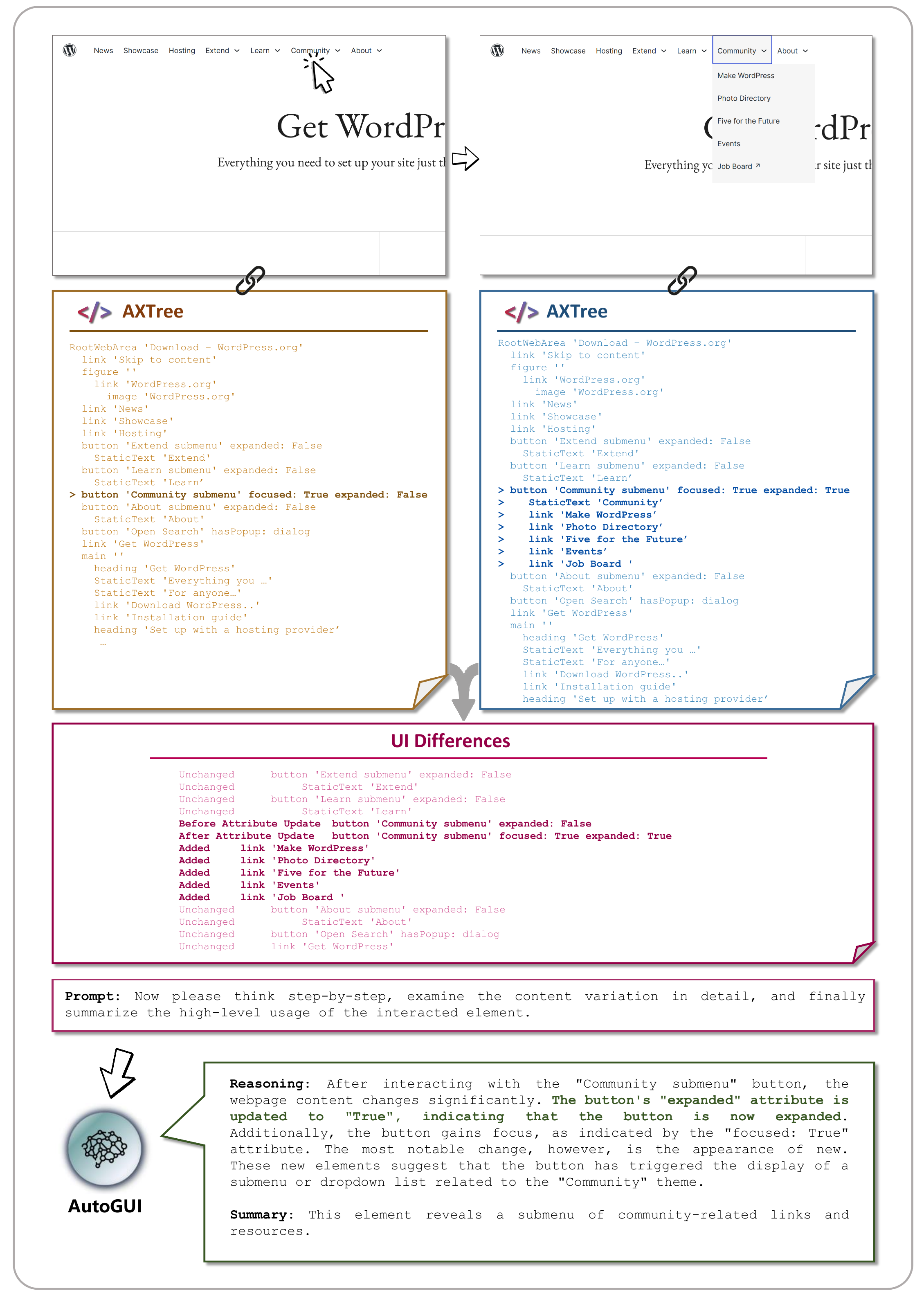}
    \caption{An example of the AutoGUI functionality annotation using UI AXTree differences. AutoGUI records the AXTrees before and after interaction and then generates line-by-line differences with our custom change markers. Subsequently, the LLM takes the differences as input to predict the element functionality.}
    \label{fig: funcpred diff case}
\end{figure*}

The AutoGUI pipeline utilizes UI content changes to predict the functionalities of the interacted elements. For interactions that manipulate the existing UI, the pipeline analyzes differences in the AXTrees to annotate functionalities. Conversely, when interactions result in navigation to a new UI, the pipeline examines changes in UI descriptions to guide the annotation process. Details on these methodologies are outlined below:

\noindent\textbf{UI manipulation case} We use a file-comparison library, DiffLib, to generate line-by-line differences of the AXtrees before and after interactions. To balance efficiency with annotation integrity, we limit the differences to 250 lines. In addition to the standard markings by DiffLib—addition, deletion, and unchanged status—we incorporate two additional change markers: `Repositioning' and `Attribute Update'. These markers provide detailed information about UI content changes, essential for representing realistic structural variations. For example, without the attribute update marker, a clicked menu icon would erroneously appear as both deleted and added in the difference output, despite the menu element remaining in place. An example of this case is shown in Fig.~\ref{fig: funcpred diff case}. The used prompt is shown in Tab.~\ref{tab:supp:funcpred manip prompt}.

\begin{table*}[tp]
\tiny
\centering
\caption{The functionality annotation prompt used in the AutoGUI pipeline in UI manipulation cases.}
\label{tab:supp:funcpred manip prompt}
\begin{tabular}{|l|}
\hline
\begin{tabular}[c]{@{}p{0.9\linewidth}@{}}\color{blue}{(Requirements for annotation)} \\
Objective: As an Internet expert, your task is to describe the usage and functionality of a webpage element based on the changes observed in the webpage contents before and after interacting with the element.\\

Instructions: \\
1. You will be shown line-by-line differences between the webpage content before and after interacting with the element. Here's what each prefix indicates:\\
Unchanged: Lines that are identical before and after the interaction.\\
Added: New lines that appear after the interaction.\\
Deleted: Lines that were present before the interaction but removed afterward.\\
Renaming: Lines indicating elements that were renamed due to the interaction.\\
Attribute Update: Lines showing elements whose attributes were updated during the interaction.\\
Repositioned: Elements that were moved to a different part of the webpage.\\
2. You MUST thoroughly analyze the changes in webpage content (Added, Deleted, Unchanged lines) caused by interacting with the element, present a detailed reasoning process elucidating how the element affects the webpage, and finally summarize the element's overall purpose based on your analysis\\
3. Avoid detailing every specific functionality of the webpage element. Instead, focus on describing its broader impact on the webpage experience. For example, if interacting with a ``Products'' button reveals a dropdown menu, do not catalog the subsequent webpage changes in exhaustive detail.\\
4. Your output MUST follow this format:\\
Reasoning: (Examine the webpage variation carefully to figure out how the interacted element changes the webpage)\\
Summary: This element ... (Provide a concise high-level description of the element's function. This description should contain the meaningful feature of the element in its context.)\\
5. Avoid mentioning specific elements from the webpage before interaction in the Summary. Instead, focus directly on the outcome of the interaction. For instance, rather than detailing a transition from one element to another, like ``This element switches the focus from the `Destination' input field to the `Check-in Date' field, providing a date picker,'' simplify it to ``This element triggers a date picker for users to select their check-in date.'' This approach keeps the description concise and focused on functionality, eliminating unnecessary details.\\
6. For simplicity, you MUST not mention the text of the element in the summary.\\
Now it's your turn.\\ \\
Element interaction: After \color{blue}{clicking a <link> element named ``Community submenu'' (an example of the interaction target)}, \color{black}{the webpage exhibits the following variations:}\\
\color{blue}{(UI content changes)}\\
Unchanged 	button `Extend submenu' expanded: False \\
Unchanged 		StaticText `Extend'\\
Unchanged 	button `Learn submenu' expanded: False\\
Unchanged 		StaticText `Learn'\\
Before Attribute Update 	button `Community submenu' expanded: False\\
After Attribute Update 	button `Community submenu' focused: True expanded: True\\
Unchanged 		StaticText `Community'\\
Added 	link `Make WordPress'\\
Added 	link `Photo Directory'\\
Added 	link `Five for the Future'\\
Added 	link `Events'\\
Added 	link `Job Board'\\
Unchanged 	button `About submenu' expanded: False\\
Unchanged 		StaticText `About'\\
Unchanged 	button `Open Search' hasPopup: dialog\\
Unchanged 	link `Get WordPress'\\ \\

Now please think step-by-step, examine the content variation in detail, and finally summarize the high-level usage of the interacted element. Do follow the output format ``Reasoning: ... Summary: This element ...''.\\ \midrule
\color{blue}{(LLM response)}\\
Reasoning: After interacting with the ``Community submenu'' button, the webpage content changes significantly. The button's ``expanded'' attribute is updated to ``True'', indicating that the button is now expanded. Additionally, the button gains focus, as indicated by the ``focused: True'' attribute. The most notable change, however, is the appearance of new. These new elements suggest that the button has triggered the display of a submenu or dropdown list related to the ``Community'' theme. \\

Summary: This element reveals a submenu of community-related links and resources.
\end{tabular} \\ \hline
\end{tabular}
\end{table*}

\begin{figure*}[th]
    \centering
    \includegraphics[width=1.0\linewidth]{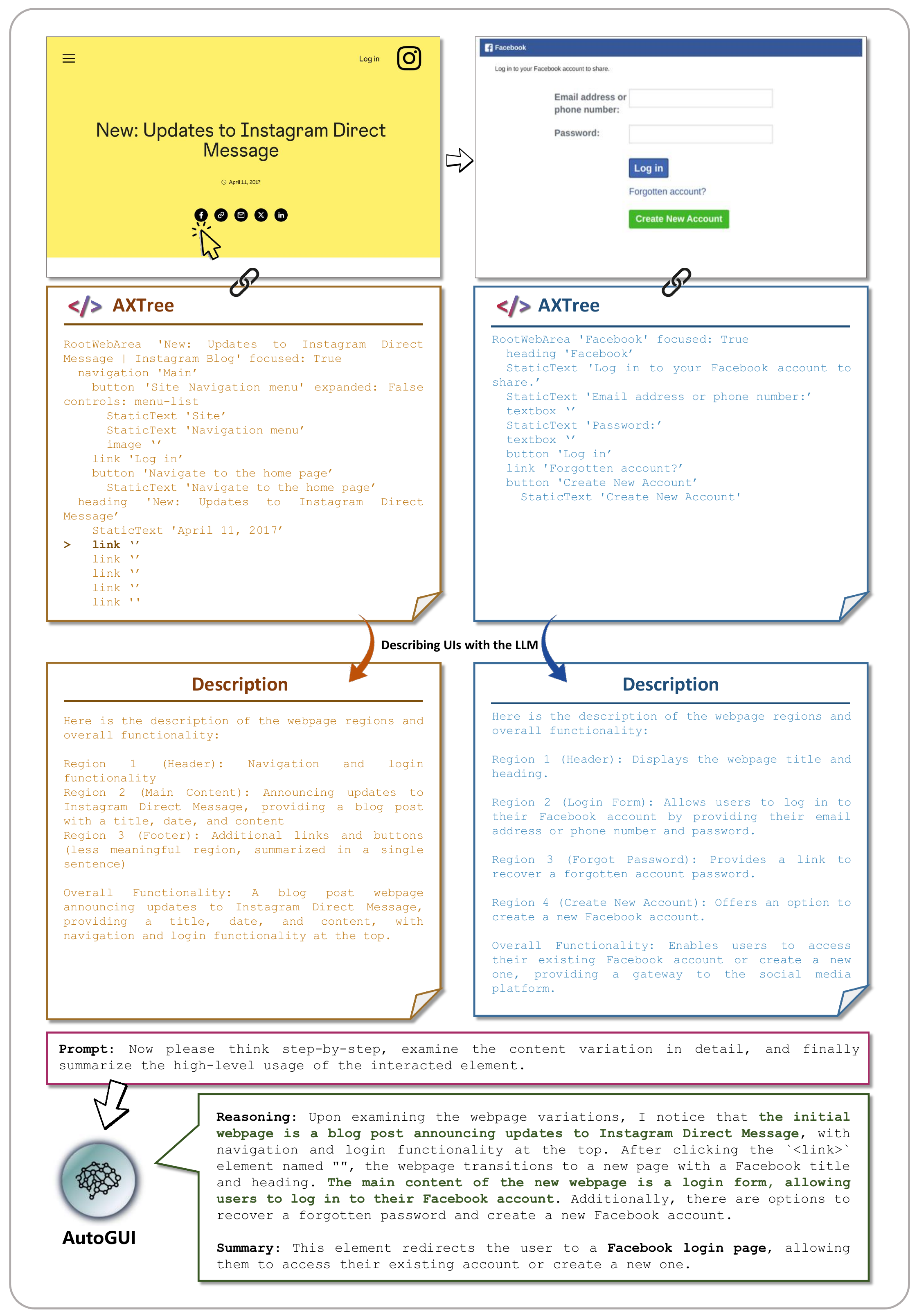}
    \caption{An example of the AutoGUI functionality annotation using UI descriptions. AutoGUI records the AXTrees before and after interaction and then prompts the LLM to describe the AXTrees in detail. Subsequently, the LLM takes the two descriptions as input to predict the element functionality.}
    \label{fig: funcpred desc case}
\end{figure*}

\noindent\textbf{UI navigation case} When an interacted element causes navigation to a new UI, the resultant changes are often extensive, potentially exceeding the context limit of an LLM and complicating the analysis of these changes. To handle this situation, UI description changes are used to predict functionalities. Concretely, the LLM is initially prompted to describe the UIs before and after interaction given UI AXTrees as input. Subsequently, the LLM then uses these descriptions to analyze content changes and predict the functionality of the interacted element. The description length of the AXTree is limited to 150 lines. An illustration of this process is shown in Fig.~\ref{fig: funcpred desc case}. The corresponding prompt is detailed in Tab.~\ref{tab:supp:funcpred nav prompt}.

\begin{table*}[tp]
\tiny
\centering
\caption{The functionality annotation prompt used in the AutoGUI pipeline in UI navigation cases. This example shows how the LLM }
\label{tab:supp:funcpred nav prompt}
\begin{tabular}{|l|}
\hline
\begin{tabular}[c]{@{}p{0.9\linewidth}@{}}\color{blue}{(Requirements for annotation)} \\
Objective: Your mission, as a digital navigation specialist, is to deduce and articulate the function and usage of a specific webpage element. This deduction should be based on your analysis of the differences in webpage content before and after interacting with said element.\\

Instructions:\\
1. You will be given descriptions of a webpage before and after interaction with an element. Your primary task is to meticulously analyze the differences in content resulting from this interaction to understand what the functionality of the element is in the webpage context.\\
2. You must present a detailed reasoning process before finally summarizing the element's overall purpose based on your analysis.\\
3. Prioritize examining changes in the webpage's regional content over individual element variations. This approach will provide a more holistic view of the element's impact on the webpage.\\
4. You should emphasize on the main content changes and pay less attention to less meaningful regions, such as headers, navigation bars, and footers.\\
5. Your output MUST follow this format:\\
Reasoning: (Examine the webpage variation carefully to figure out how the interacted element changes the webpage)\\
Summary: This element ... (Provide a high-level description of the element's functionality. This description should contain the meaningful feature of the element in its context.)\\
6. Avoid mentioning specific elements from the webpage prior to interaction in the Summary. Instead, focus directly on the outcome of the interaction. For instance, rather than detailing element changes, like ``This element triggers the disappearance of the header and language selector elements and the emergence of a login form'', simplify it to ``This element triggers the display of a login page.'' This approach keeps the summary concise and focused on functionality, eliminating unnecessary details.\\
7. The final summary must emphasize the distinctive functionality of the element and not be too vague. For example, ``This element redirects the user to a specific website, allowing them to access its content and services''  is too broad and lacks meaningful detail. Instead, it should precisely describe how the element provides users with unique value\\

Now it's your turn.\\
The description of the initial webpage:\\
\color{blue}{(An example description of the UI before the interaction)}\\
Region 1 (Header): Navigation and login functionality\\
Region 2 (Main Content): Announcing updates to Instagram Direct Message, providing a blog post with a title, date, and content\\
Region 3 (Footer): Additional links and buttons (less meaningful region, summarized in a single sentence)\\
Overall Functionality: A blog post webpage announcing updates to Instagram Direct Message, providing a title, date, and content, with navigation and login functionality at the top.\\

After \color{blue}{clicking a <link> element named ``'' (an example of the interaction target)}, we jump to a new webpage.\\
The description of the new webpage is:\\
\color{blue}{(An example description of the UI after the interaction)}\\
Region 1 (Header): Displays the webpage title and heading.\\
Region 2 (Login Form): Allows users to log in to their Facebook account by providing their email address or phone number and password.\\
Region 3 (Forgot Password): Provides a link to recover a forgotten account password.\\
Region 4 (Create New Account): Offers an option to create a new Facebook account.\\
Overall Functionality: Enables users to access their existing Facebook account or create a new one, providing a gateway to the social media platform.\\ \\

Now please think step-by-step, examine the content variation in detail, and finally summarize the high-level usage of the interacted element. Do follow the output format ``Reasoning: ... Summary: This element ...''.\\ \midrule
\color{blue}{(LLM response)}\\
Reasoning: Upon examining the webpage variations, I notice that the initial webpage is a blog post announcing updates to Instagram Direct Message, with navigation and login functionality at the top. After clicking the `<link>` element named ``'', the webpage transitions to a new page with a Facebook title and heading. The main content of the new webpage is a login form, allowing users to log in to their Facebook account. Additionally, there are options to recover a forgotten password and create a new Facebook account.\\
Summary: This element redirects the user to a Facebook login page, allowing them to access their existing account or create a new one. \\

\end{tabular} \\ \hline
\end{tabular}
\end{table*}

\subsubsection{Details of Rejecting Invalid Samples via Hand-Written Rules}
\label{sec:supp:hand-written rules}
To clarify the hand-written rules used in the process of removing invalid samples: (1) \textbf{Removing blank GUIs}. We remove blank GUIs by verifying whether the accessibility tree contains more than one node. If no nodes are present, the GUI is considered blank. (2) \textbf{Removing UIs containing elements indicating content loading}. GUIs containing elements indicative of content loading (e.g., keywords such as "loading", "please wait", or "refreshing") are excluded. These keywords typically suggest that the content has not fully loaded and may affect the validity of the sample. (3) \textbf{Removing interaction targets outside of screens}. Occasionally, part of the UI, including the interacted element, may fail to be captured. We filter out GUIs where interaction targets appear outside of the visible screen area. This is determined by checking whether the interacted element exists within the bounds of the recorded accessibility tree. Note that these rules are designed mainly for the domains from which we collected GUI metadata. Nevertheless, one can extend the rules flexibly according to the noise characteristics of new domains.

\subsubsection{Details of Rejecting Invalid Samples via LLMs}
\label{suc:supp:reject details}
\noindent\textbf{Rejection process} To eliminate invalid samples before functionality annotation, the AutoGUI pipeline prompts the annotating LLM to also determine the validity of samples by analyzing the predictability of the UI content changes. The LLM evaluates each sample against three criteria: 1) Explicitness of Changes: This measures how clearly the changes indicate the element's functionality. Changes that directly suggest functionality receive higher scores, while vague or irrelevant changes are not scored. 2) Relevance of Changes: This criterion assesses the significance of the modifications in relation to the element's intended function. Highly related modifications obtain a high score. No scores for irrelevant or unrelated content changes. 3) Predictability of Outcome: This involves determining how anticipated the interaction outcome is based on the changes, considering common web conventions and user experience principles. Highly predictable changes obtain a high score, whereas moderate, unexpected, or counter-intuitive outcomes receive no score.

\subsubsection{Details of LLM-Based Verification}
\label{suc:supp:verif details}
\begin{figure*}[th]
    \centering
    \includegraphics[width=1.0\linewidth]{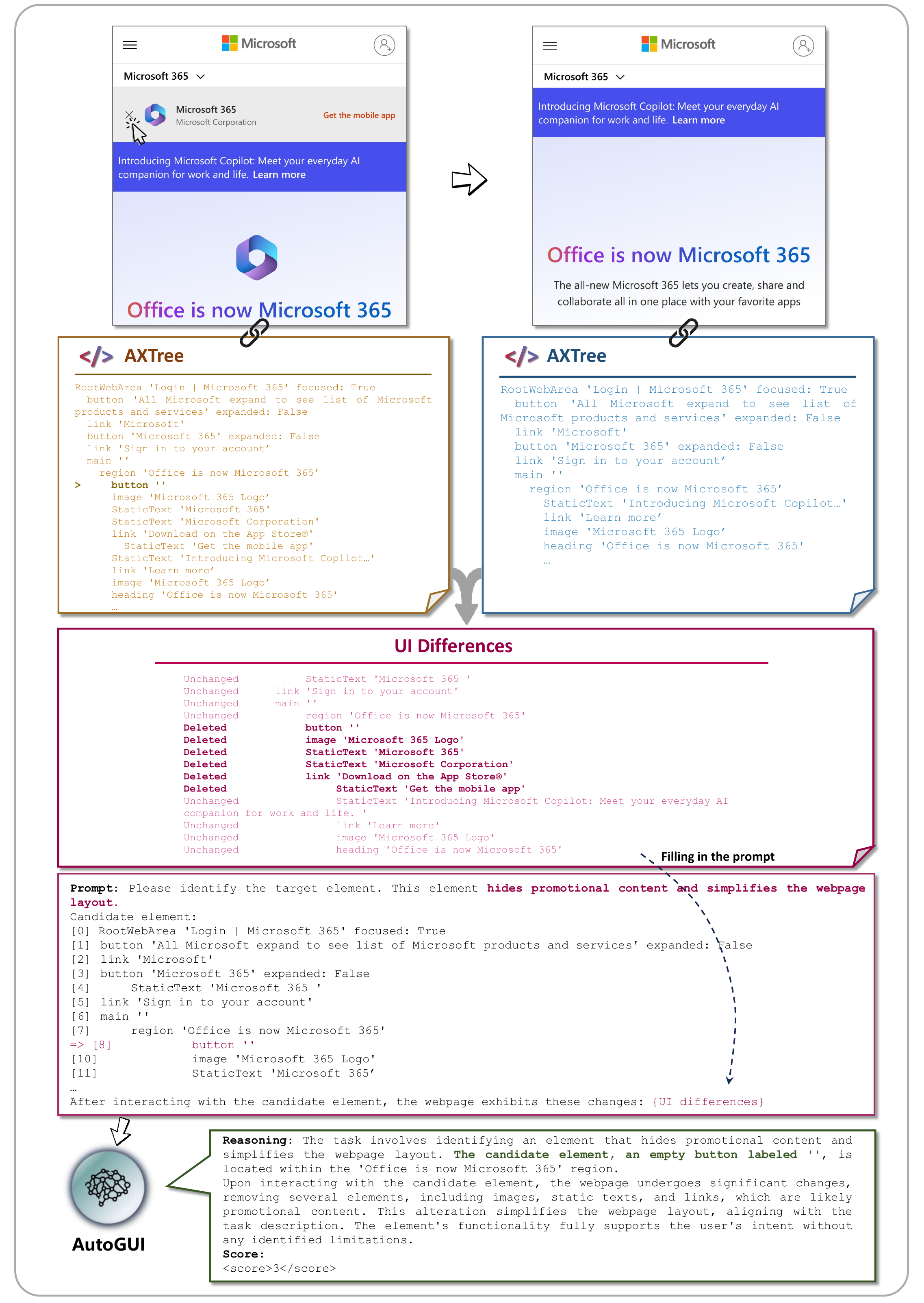}
    \caption{An example of AutoGUI prompting the LLM as a self-verifier to determine whether an element supports its functionality annotation. The sample shown is a manipulation case in which AutoGUI uses UI line-by-line differences to make decisions about whether a button fulfills the intent of hiding promotional content.}
    \label{fig: verif diff case}
\end{figure*}

\noindent\textbf{Verification process} To improve the quality of functionality annotations, the AutoGUI pipeline prompts two LLMs (i.e.g, Llama-3-70B and Mistral-7B-Instruct-v0.2) as verifiers to assign scores to samples based on how well the target elements adhere to their functionality annotations. The LLMs receive as the input a) the target element along with its surrounding UI content (up to 20 lines), b) the functionality annotation of this element, and c) the outcome of interacting with the element, either being the UI line-by-line differences (at most 250 lines) in manipulation cases or the UI description after the interaction in navigation cases. Given these inputs, the two LLMs generate two responses containing a score. Samples that do not achieve two full scores are discarded for higher quality of the AutoGUI dataset. The used prompt is shown in Tab.~\ref{tab:supp:verif prompt} and an example is illustrated in Fig.~\ref{fig: verif diff case}.

\begin{table*}[tp] 
\tiny
\centering
\caption{The rejection prompt used in the AutoGUI pipeline in UI manipulation cases. This example shows how the LLM assigns a low score to a sample that exhibits meaningless and unpredictable UI content changes.}
\label{tab:supp:rejection prompt}
\begin{tabular}{|l|}
\hline
\begin{tabular}[c]{@{}p{0.9\linewidth}@{}}\color{blue}{(Requirements for rejection)} \\Your primary objective is to determine whether the changes in the webpage's content are sufficient for predicting the functionality of the webpage element causing these changes after being interacted with.\\

Instructions:\\
1. You will be shown the outcome (webpage changes) resulting from interacting with the element. The outcome can take one of two forms:  changes to the webpage description, or line-by-line differences. For the latter form, here's what each prefix indicates:\\
Unchanged: Lines that are identical before and after the interaction.\\
Added: New lines that appear after the interaction.\\
Deleted: Lines that were present before the interaction but removed afterward.\\
Renaming: Lines indicating elements that were renamed due to the interaction.\\
Attribute Update: Lines showing elements whose attributes were updated during the interaction.\\
Repositioned: Elements that were moved to a different part of the webpage.\\
2. Analyze the provided outcome and provide detailed reasoning for whether this outcome helps to predict the element's functionality, considering the following stringent criteria:\\
1) Explicitness of Changes: Rate how directly the changes suggest the element's functionality. Score 1-3 for clear, unambiguous changes. Clearer changes obtain a higher score. No scores for vague, meaningless, or non-specific changes.\\
Positive Example: A button labeled ``Show More'' that, upon interaction, clearly adds new content below it. The direct addition of content clearly indicates a content expansion functionality. Score: 3\\
Negative Example: After clicking a ``Details'' button, the page layout changes subtly without adding relevant information or altering content in a meaningful way. The changes do not clearly relate to the button's presumed functionality. Score: 0\\

2) Relevance of Changes: Evaluate the significance of the modifications in relation to the element's intended function. Score 1-3 for changes that enhance understanding of the element’s role. Highly related modifications obtain a high score. No scores for irrelevant or unrelated content changes.\\
Positive Example: Clicking on a ``Contact Us'' button opens a form to fill out, which is highly relevant to the button's intended functionality. Score: 3\\
Negative Example: Clicking on a ``View Profile'' link leads to a page refresh without displaying the profile or any related information, making the change irrelevant to the link's intended purpose. Score: 0\\

3) Predictability of Outcome: Assess how anticipated the interaction outcome is based on the changes, considering common web conventions and user experience principles. Score 1-3 for highly predictable outcomes. Highly predictable changes obtain a high score. No scores for outcomes that are moderate, unexpected, or counterintuitive.\\
Positive Example: Clicking or hovering over a ``Help'' icon reveals a tooltip with information, an outcome that users can easily predict given the icon's universal symbol for help. Score: 3\\
Negative Example: Hovering over a ``Settings'' icon changes its color but does not display any settings options, tooltips, or lead to a settings page, making the outcome unpredictable and the icon's purpose unclear. Score: 0\\

Provide a detailed analysis based on these criteria, ending with an overall score between 0-9.\\

4. Your output MUST follow this format:\\
Reasoning:\\
- Explicitness of Changes: [Detailed reasoning and score]\\
- Relevance of Changes: [Detailed reasoning and score]\\
- Predictability of Outcome: [Detailed reasoning and score]\\

Overall score: <score>[criterium 1 score] + [criterium 2 score] + [criterium 3 score] = [total score]</score> (output an additive equation enclosed with a pair of <score></score>)\\

Now it's your turn.\\
The interacted element is \color{blue}{a <button> element named ``Cancel'' (interaction target)}\\
Outcome:\\
\color{blue}{(An example of interaction outcome)}\\
After clicking a <button> element named ``Cancel'', the webpage exhibits the following variations:\\
Unchanged 			StaticText `Email Address'\\
Unchanged 			textbox `Email Address' required: True\\
Unchanged 			button `Submit'\\
Unchanged 				StaticText `Submit'\\
Before Attribute Update 			button `Cancel'\\
After Attribute Update 			button `Cancel' focused: True\\
Unchanged 				StaticText `Cancel'\\
Unchanged 	contentinfo `'\\
Unchanged 		StaticText `Your contact information is used to deliver requested updates or to access your subscriber preferences.'\\ \\

Now please think step-by-step, examine the content variation in detail, and finally calculate scores according to the criteria. Do follow the output format ``Reasoning: ... Overall score: <score>...</score>''.\\ \midrule
\color{blue}{(LLM response)}\\
Reasoning:\\

- Explicitness of Changes: The changes resulting from interacting with the ``Cancel'' button are mostly unchanged elements, except for the ``Cancel'' button itself, which now has a ``focused: True'' attribute. This change is not directly related to the button's functionality, as it does not provide clear insight into what the button does when clicked. Score: 1\\

- Relevance of Changes: The changes do not provide any significant information about the button's intended function. The button's name ``Cancel'' suggests that it might cancel some action or process, but the changes do not reveal what exactly is being canceled or how the button's functionality relates to the surrounding elements. Score: 1\\

- Predictability of Outcome: Given the name ``Cancel'' and the context of a form with an ``Email Address'' field and a ``Submit'' button, a user might expect the ``Cancel'' button to reset the form, clear the input field, or abort a submission process. However, the outcome does not provide any clear indication of what happens when the button is clicked. The changes are not predictable based on common web conventions and user experience principles. Score: 1\\

Overall score: <score>1 + 1 + 1 = 3</score>

\end{tabular} \\ \hline
\end{tabular}
\end{table*}
\begin{figure*}[th]
    \centering
    \includegraphics[width=1.0\linewidth]{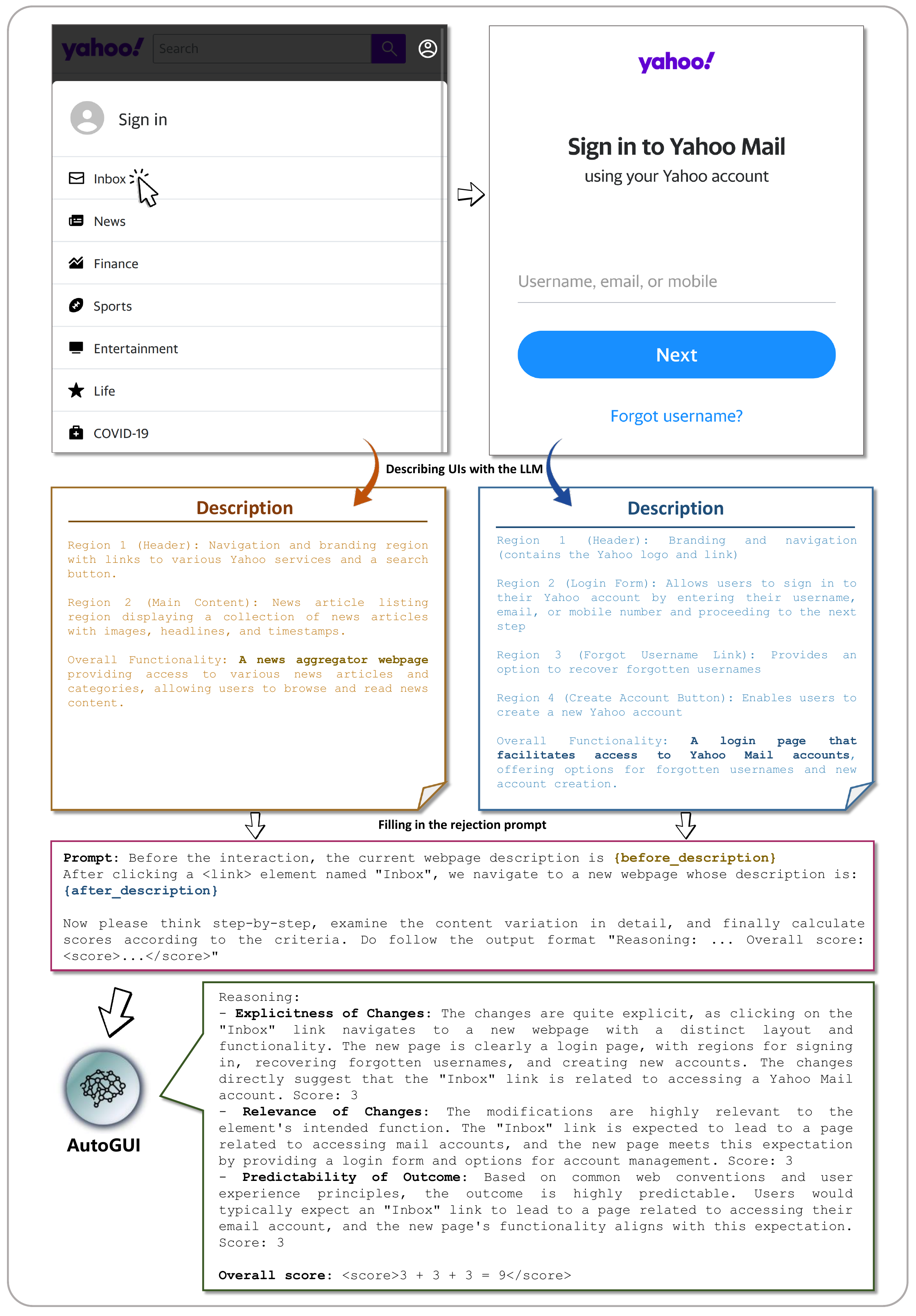}
    \caption{An example of AutoGUI prompting the LLM as a rejector to determine whether a sample shows meaningful UI content changes sufficient for predicting the functionality of the interacted element. The sample shown is a navigation case in which AutoGUI uses UI descriptions, instead of line-by-line differences, to make decisions.}
    \label{fig: rejection desc case}
\end{figure*}

\begin{figure*}[th]
    \centering
    \includegraphics[width=1.0\linewidth]{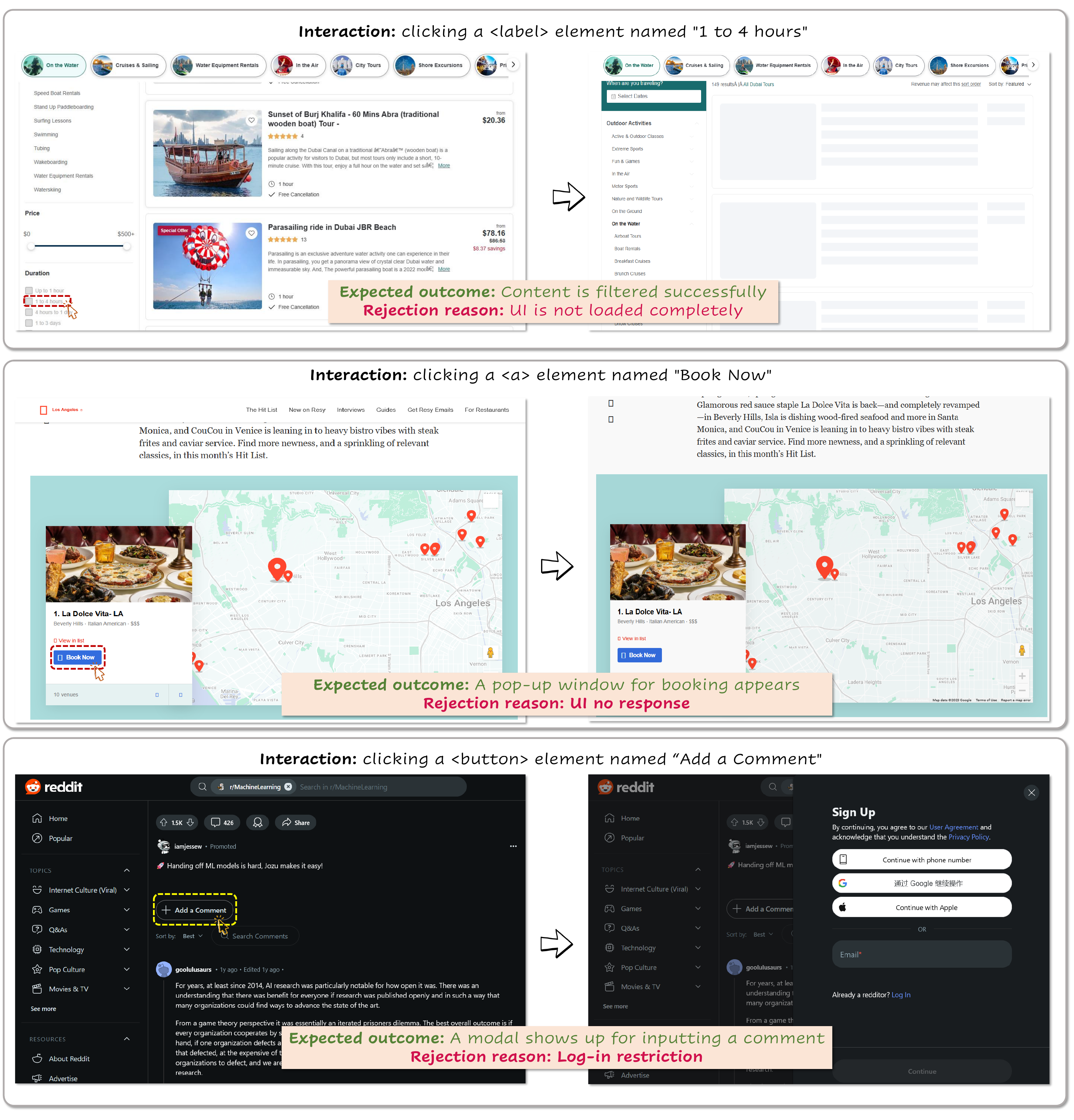}
    \caption{\textbf{Examples of samples rejected by the AutoGUI pipeline.} The first sample encounters incompletely loaded content that interferes LLM annotation. The second encounters a no-response issue where the pop-up window fails to appear. The third shows a case where an unexpected log-in page pops up to interrupt the functionality of the ``Add a Comment'' element.}
    \label{fig: rejection examples}
\end{figure*}

\begin{table*}[tp] 
\tiny
\centering
\caption{The self-verification prompt used in the AutoGUI pipeline in UI manipulation cases. This example shows how the LLM assigns a low score to the incorrect functionality.}
\label{tab:supp:verif prompt}
\begin{tabular}{|l|}
\hline
\begin{tabular}[c]{@{}p{0.9\linewidth}@{}}
\color{blue}{(Requirements for self-verification)}\\
Given the following inputs:\\
1) Webpage content: This input represents the hierarchical structure of a webpage's elements, emphasizing semantic information and relationships. Each node in the tree includes details such as the element's role (e.g., button, link, heading), relevant attributes (e.g., expanded), and hierarchical relationships with other elements.\\

2) Task Description: This describes the action a user intends to perform (such as submitting a form, navigating to a particular section, or adjusting settings) or the information they seek (such as a specific content piece or form field). It also introduces a candidate element for evaluation and then presents the webpage changes caused by interacting with this element. Your task is to assess whether this element effectively facilitates the specified user action.\\

Your job is to:\\
1) Analyze the provided webpage content to understand the structure and semantics of the webpage's elements.\\
2) Evaluate the Candidate Element: Determine the suitability of the specified candidate element for the described action. Consider the element's role, attributes, and position within the hierarchy. Your evaluation should be grounded in how well these aspects align with the required functionality for the user's intended action.\\
3) Score the Element: Assign a score ranging from 0 to 3, enclosed within <score></score> tags. This score should reflect the degree to which the candidate element meets the action's requirements:\\
0: The element does not support the action in any capacity.\\
1: The element provides minimal support for the action.\\
2: The element supports the action but with limitations.\\
3: The element fully supports the action without significant limitations.\\
4) Provide Reasoning: Before presenting your score, offer a detailed explanation of your reasoning. This should cover your analysis of the webpage content, the relationship between the candidate element and the specified action, and how these factors informed your scoring decision.\\
5) Format for Your Answer:\\
Reasoning: (Provide a comprehensive analysis covering the webpage's insights, the relationship between the specified action and the candidate element, and the rationale behind your scoring decision.)\\

Score: <score>$[Your Score Here]$</score>

\color{blue}{(An in-context exemplar)}\\
Example:\\
Webpage content:\\
$[0]$ RootWebArea `Rental Cars at Low, Affordable Rates'\\
$[1]$ dialog `Vehicle Class' modal: True\\
$[2]$ radiogroup `Vans' \\
$[3]$ 	radio `Minivans' checked: false\\
$[4]$ 	radio `Passenger Vans' checked: false\\
$[5]$ 	radio `Cargo Vans' checked: false\\
$[6]$ button `Cancel'\\
$[7]$ button `Apply Filter'\\

Task Description: Please identify the target element. The element helps users narrow down their vehicle choices to minivans specifically\\

Candidate element: $[3]$ 	radio `Minivans'\\
After interacting with the candidate element, the webpage exhibits these changes:\\
Upon clicking the ``Minivans'' <input> element, a new ``Remove Filter'' button is added to the dialog modal. The radio buttons for different vehicle classes remain unchanged, but the ``Minivans'' radio button is checked after the interaction.\\

Reference response:\\
Reasoning: The provided webpage content outlines a clear hierarchical structure for selecting vehicle types on a car rental webpage, categorized into sections like Cars, Vans, and Trucks, each with its own set of options represented as radio buttons. \\
The task involves narrowing vehicle choices to minivans. The candidate element is part of the `Vans' radiogroup on a car rental webpage. This directly supports the user's action of narrowing choices to minivans. The element's interaction leads to its checking and introduces a ``Remove Filter'' button. Its role, position, and functionality support the user's intent.\\

Score: <score>3</score>

\color{blue}{(The self-verification task)}\\
Now it's your turn.\\
Current webpage content:
\color{blue}{(A UI content example)} \\
$[0]$ RootWebArea `X for Business | X tips, tools, and best practices' focused: True \\
$[2]$ 			link `Get started in your ads account'\\
$[4]$ 			link `Campaign setup'\\
$[6]$ 			link `Editing and optimization'\\
$[8]$ 			link `Measurement'\\
$[9]$ 	heading ``Be what's happening''\\
$[10]$ 	StaticText `Spread your message, grow your audience, and get more traffic with Twitter Ads.'\\
=> $[11]$ 	link `Learn more' \color{blue}{(The interaction target)}\\
$[13]$ 	heading `Twitter is where people come to discover what’s happening.'\\
$[14]$ 	StaticText `With over 500 million Tweets being sent each day, ...'\\
$[15]$ 	StaticText `Whether you’re a CEO strengthening their personal brand...'\\
$[16]$ 	link `Get started with Twitter for Business'\\
$[17]$ 		button `Get started with Twitter for Business'\\
$[18]$ 	heading ``What's new on Twitter Business Outlook''\\
$[20]$ 	link `Google'\\

Task Description: Please identify the target element. This element \color{blue}{redirects users to a non-existent page, providing a custom error message and alternative content instead of the expected information. (The functionality annotation of the element)}\\

Candidate element: \color{blue}{$[11]$ 	link `Learn more'  (The interaction target)}\\
\color{blue}{(The interaction outcome)}\\
After interacting with the candidate element, we navigate to a new webpage that contains these contents:\\
Region 1 (Header): Contains the website's logo and navigation links.\\
Region 2 (Error Message): Displays an error message indicating that the page does not exist, accompanied by a humorous image and a link to potentially related content.\\
Region 3 (Footer): Provides links to various policy and information pages, as well as copyright information.\\
Overall Functionality: This webpage serves as a custom 404 error page, providing a lighthearted and engaging way to notify users that the requested page is not available.\\ \\

Now please do your job. Your answer: \\ \midrule
\color{blue}{(LLM response)}\\
Reasoning: The candidate element, a link labeled `Learn more', is part of the webpage's main content, surrounded by descriptions and headings related to Twitter for Business. However, upon interacting with this element, the webpage exhibits significant changes, redirecting users to a custom 404 error page with a humorous image and a link to potentially related content. In fact, the element's interaction leads to a completely different webpage with an error message and alternative content, which is unrelated to the initial webpage's purpose. The element's role, position, and functionality do not align with the user's intended action.\\

Score: <score>0</score>
\end{tabular} \\ \hline
\end{tabular}
\end{table*}

\subsubsection{Details of Grounding/Captioning Task Generation}
After collecting the element-functionality pairs, the AutoGUI pipeline converts these pairs into functionality grounding and captioning tasks. A functionality grounding task requires a VLM to output point coordinates of the element fulfilling the given functionality, while a captioning task demands that the VLM articulate a functionality description for an element, given its coordinates. It is important to note that each element-functionality pair is utilized to generate both a grounding task and a captioning task.

\subsection{Implementation Details}
\subsubsection{Human Evaluation Details}
\label{sec:supp:humaneval details}
To justify the efficacy of the AutoGUI pipeline, we conducted a comparative evaluation of annotation correctness between a trained human annotator and the AutoGUI system. The human annotator was a graduate student proficient in using digital devices, ensuring familiarity with diverse user interfaces.

We selected a set of 30 invalid samples, each showcasing a variety of element functionalities, to prepare the annotator for the annotation process. These functionalities included drop-down menu expansions, menu item selections, date-pickers, filtering options, pop-up modals, webpage navigation, and zooming in/out buttons. The purpose of this selection was to expose the annotator to a broad spectrum of potential UI interactions, enhancing their ability to accurately assess element functionality based on UI content changes.

\begin{figure*}[t]
    \centering
    \includegraphics[width=0.95\linewidth]{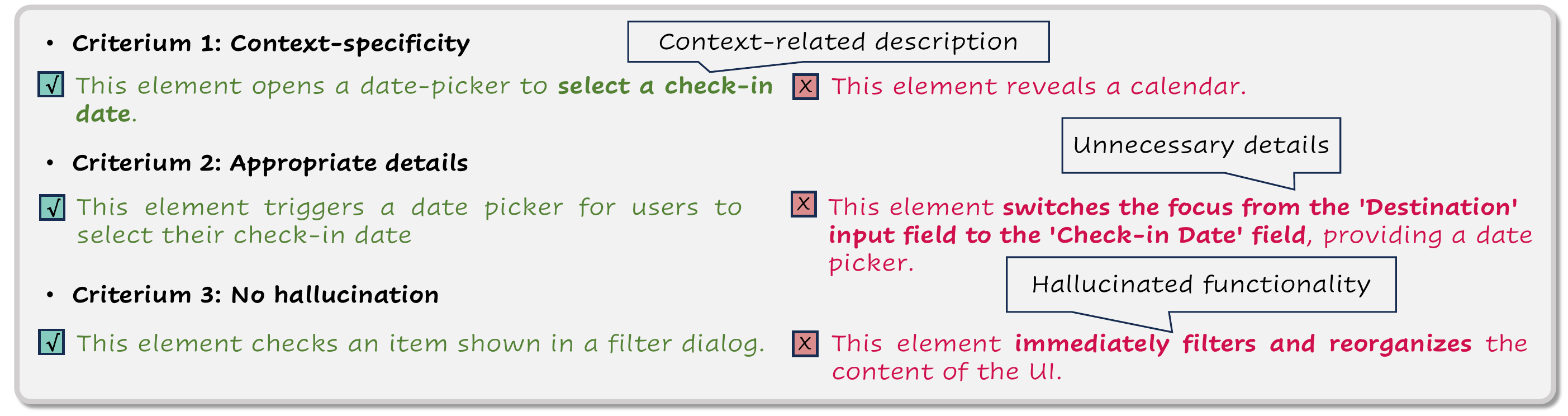}
    \caption{The checking criteria used for comparing AutoGUI pipeline and the human annotator.}
    \label{fig: check criteria}
\end{figure*}

During the training phase, we provided the annotator with detailed guidelines, including three specific criteria outlined in Fig~\ref{fig: check criteria}, to ensure the clarity and correctness of their annotations. Additionally, we incorporated 15 invalid samples to instruct the human annotator on how to identify and exclude these cases during the evaluation process. These invalid samples encompassed scenarios such as incompletely loaded UIs, network failure incidents, login restrictions, and UIs displaying inappropriate content.

Following the training stage, the human annotator evaluated a total of 146 samples. Remarkably, the annotator successfully identified all invalid samples, achieving an overall annotation correctness rate of 95.5\%. The few incorrect annotations were categorized as such due to vagueness or instances of hallucination, where the descriptions did not accurately reflect the UI elements.

\subsubsection{Fine-Tuning Details}
\label{sec:supp:impl details}
Qwen-VL-Chat~\cite{bai2023qwen}, SliME~\cite{slime}, and Qwen2-VL-7B~\citep{qwen2vl} are selected as the base models in the experiments. To investigate the scaling effects of our dataset, 25k, 125k, and the entirety of the 702k samples in the training split are used as training data in the three scaling experiments. For the first two smaller-scale experiments, a subset of the 702k data is randomly sampled.

Pilot experiments find that the non-UI training data (i.e., LLaVA-instruct-150k and the Cauldron) significantly outnumber the 25k and 125k UI training data, resulting in data imbalance that biases the trained UI-VLM towards the general Q\&A tasks in the non-UI data and leads to inferior UI grounding performance. To tackle this issue, the 25k/125k samples are resampled to the same number of the non-UI training data to enable the UI-VLM to acquire more supervising signals from the UI data. This resampling approach is not employed in the 702k experiment as this experiment does not encounter the imbalance issue.

\begin{table}[t]
\centering
\small
\caption{The training hyper-parameters used for fine-tuning Qwen-VL in the experiments.}
\label{tab:training cfg qwen}
\begin{tabular}{@{}cc@{}}
\toprule
Hyper-Parameter & Value \\ \midrule
Epoch & 1 \\
Global batch size & 128 \\
\#GPUs & 8 \\
LoRA Rank & 64\\
LoRA Alpha & 16\\
Learning rate & 3e-5 \\
weight decay & 0.1 \\
ADAM Beta2 & 0.95 \\
Warm-up ratio & 0.01 \\
LR scheduler & Cosine \\
Model max length & 768 \\
LoRA & ViT + LLM \\
DeepSpeed & ZeRO-2 \\
\#Parameters & \begin{tabular}[c]{@{}c@{}}Trainable params: 234,500,864\\ All params: 9,891,436,032\\ Trainable\%: 2.3707\end{tabular} \\
Data type & BFloat16 \\ \bottomrule
\end{tabular}
\end{table}
\begin{table}[t]
\centering
\small
\caption{The training hyper-parameters used for fine-tuning SliME in the experiments.}
\label{tab:training cfg slime}
\begin{tabular}{@{}cc@{}}
\toprule
Hyper-Parameter & Value \\ \midrule
Epoch & 1 \\
Global batch size & 128 \\
\#GPUs & 8 \\
Learning rate & 3e-5 \\
weight decay & 0.0 \\
ADAM Beta2 & 0.95 \\
Warm-up ratio & 0.03 \\
LR scheduler & Cosine \\
Model max length & 2048 \\
Frozen module & ViT \\
DeepSpeed & ZeRO-2 \\
\#Parameters & \begin{tabular}[c]{@{}c@{}}Trainable params: 7535796224\\ All params: 8364644352\\ Trainable\%: 90.09\end{tabular} \\
Data type & BFloat16 \\ \bottomrule
\end{tabular}
\end{table}
\begin{table}[t]
\centering
\small
\caption{The training hyper-parameters used for fine-tuning Qwen2-VL in the experiments.}
\label{tab:training cfg qwen2vl}
\begin{tabular}{@{}cc@{}}
\toprule
Hyper-Parameter & Value \\ \midrule
Epoch & 1 \\
Global batch size & 128 \\
\#GPUs & 8 \\
LoRA Rank & 128 \\
LoRA Alpha & 256 \\
Learning rate & 3e-5 \\
weight decay & 0.0 \\
ADAM Beta2 & 0.95 \\
Warm-up ratio & 0.03 \\
LR scheduler & Cosine \\
Model max length & 2048 \\
Frozen module & ViT \\
DeepSpeed & ZeRO-0 \\
\#Parameters & \begin{tabular}[c]{@{}c@{}}Trainable params: 322,961,408\\ All params: 8,614,337,024\\ Trainable\%: 3.75\end{tabular} \\
Data type & BFloat16 \\ \bottomrule
\end{tabular}
\end{table}

We train our models based on the HuggingFace Transformers\footnote{https://huggingface.co/docs/transformers/index} and the PEFT library\footnote{https://huggingface.co/docs/peft/index}. The training configurations are shown in Tab.~\ref{tab:training cfg qwen}, Tab.~\ref{tab:training cfg slime}, and Tab.~\ref{tab:training cfg qwen2vl}.

Fine-tuning Qwen-VL-AutoGUI702k, SLiME-AutoGUI702k, Qwen2-VL-7B-AutoGUI702k, SeeClick w/ AutoGUI702k, UGround w/ AutoGUI702k consumed approximately 25 hours, 36 hours, 20 hours, 25 hours, 46 hours, respectively.

The framework used to fine-tune Qwen-VL and SeeClick w/ AutoGUI702k is the SeeClick codebase~\citep{cheng2024seeclick}; The framework used to fine-tune SLiME-AutoGUI702k is the SLiME codebase~\citep{slime}; The framework used to fine-tune Qwen2-VL-7B-AutoGUI702k and UGround w/ AutoGUI702k is LLaMA-Factory~\citep{zheng2024llamafactory}. 

\subsubsection{Samples of Benchmarks}
\begin{figure*}[t]
    \centering
    \includegraphics[width=0.95\linewidth]{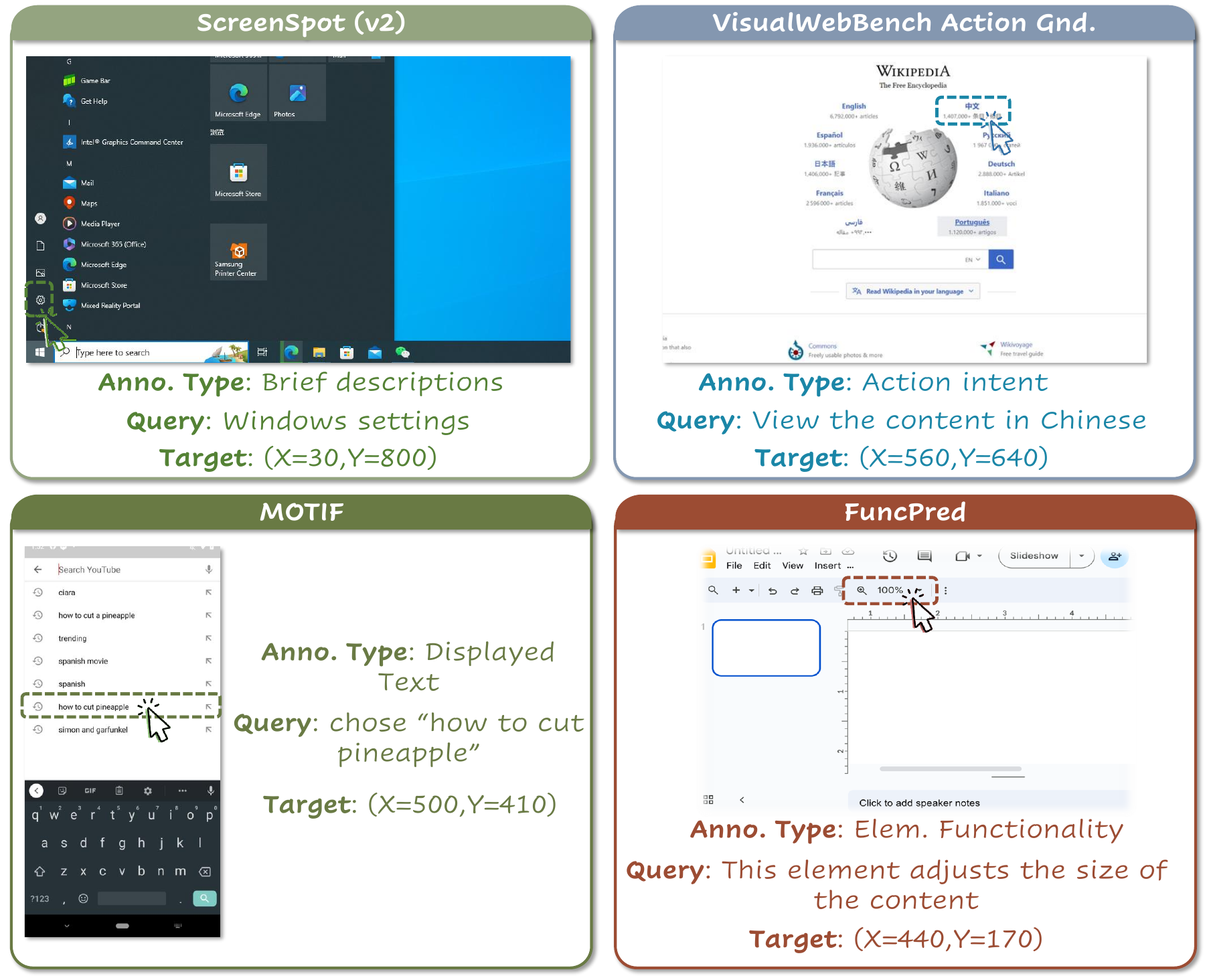}
    \caption{Samples of the UI grounding benchmarks used in the experiments}
    \label{fig: bmk samples}
\end{figure*}

For clarity, the benchmarks' samples are visualized in Fig.~\ref{fig: bmk samples}.

\subsection{Details for Potential Use of AutoGUI Dataset}
\label{sec:supp:2stage planning}
We mainly conduct 2-stage planning on the AITW~\citep{rawles2023android} benchmark to assess the benefits of our AutoGUI data on downstream agent tasks.

As illustrated in Fig.~\ref{fig: 2stage example}, a planner is utilized to conduct reasoning and step prediction while a grounding model locates target elements for the actions that require targets (click, long-press, and hover). For other actions like swipe, back, home, and input-text, the grounding model is not involved. As this experiment requires the planner to describe the expected functionality of target elements, we use strong proprietary VLMs, such as GPT-4o-mini and Gemini-2.0 as the planners. Expert models, such as OS-ATLAS~\citep{osatlas}, are not used as they typically have lost general instruction following capability after large-scale fine-tuning.

The results in Tab.~\ref{tab:2-stage planning} show that Qwen2-VL-7B trained with our functionality grounding tasks can help the planners to more accurately locate target elements.

\subsection{Additional Experimental Analysis}
\subsubsection{Growing Grounding Performance Brought by Scaling Data Size}
\begin{figure*}[h]
    \centering
    \includegraphics[width=1.0\linewidth]{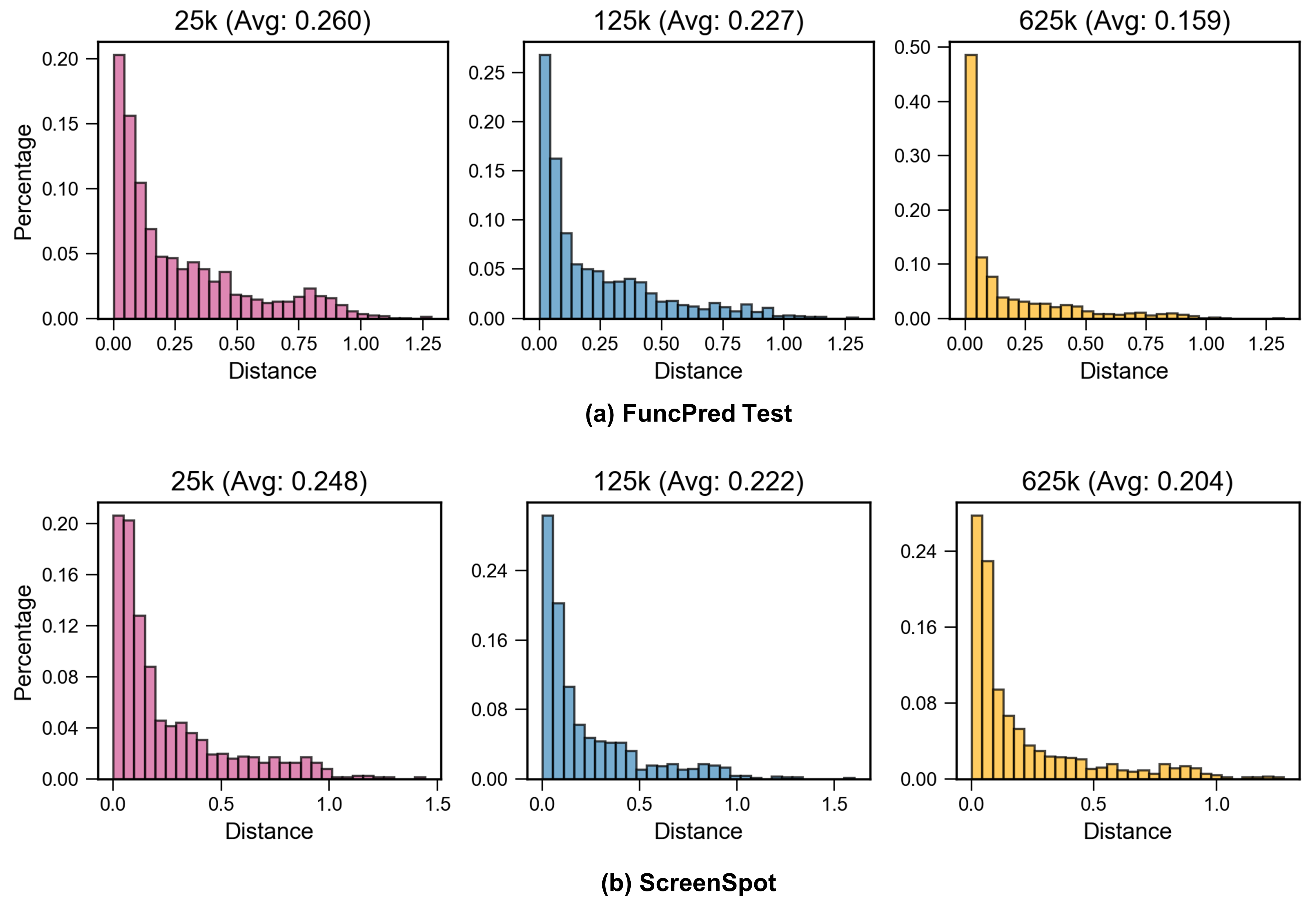}
    \caption{\textbf{Histograms of distances from predicted points to ground truth box centers.} The distance from the normalized coordinate of a predicted point to its corresponding GT box center is calculated for all samples. Then, the histograms of these distances are illustrated to demonstrate the growing grounding performances brought by scaling the AutoGUI data size. The averaged distance for each experiment is displayed on the subplot title.}
    \label{fig: dist histogram}
\end{figure*}

To further investigate the benefit of scaling the AutoGUI functionality data, the histogram of distance from a predicted point to the ground truth box center is plotted for the 25k, 125k, and 702k experiments. The results in Fig.~\ref{fig: dist histogram} demonstrate that the distance distributions become denser at lower ranges, suggesting that increasing the AutoGUI training data leads to consistently improved grounding performances.

\subsubsection{Case Analysis on FuncPred Test Split}
\label{sec:supp:case analysis}
\begin{figure*}[th]
    \centering
    \includegraphics[width=1.0\linewidth]{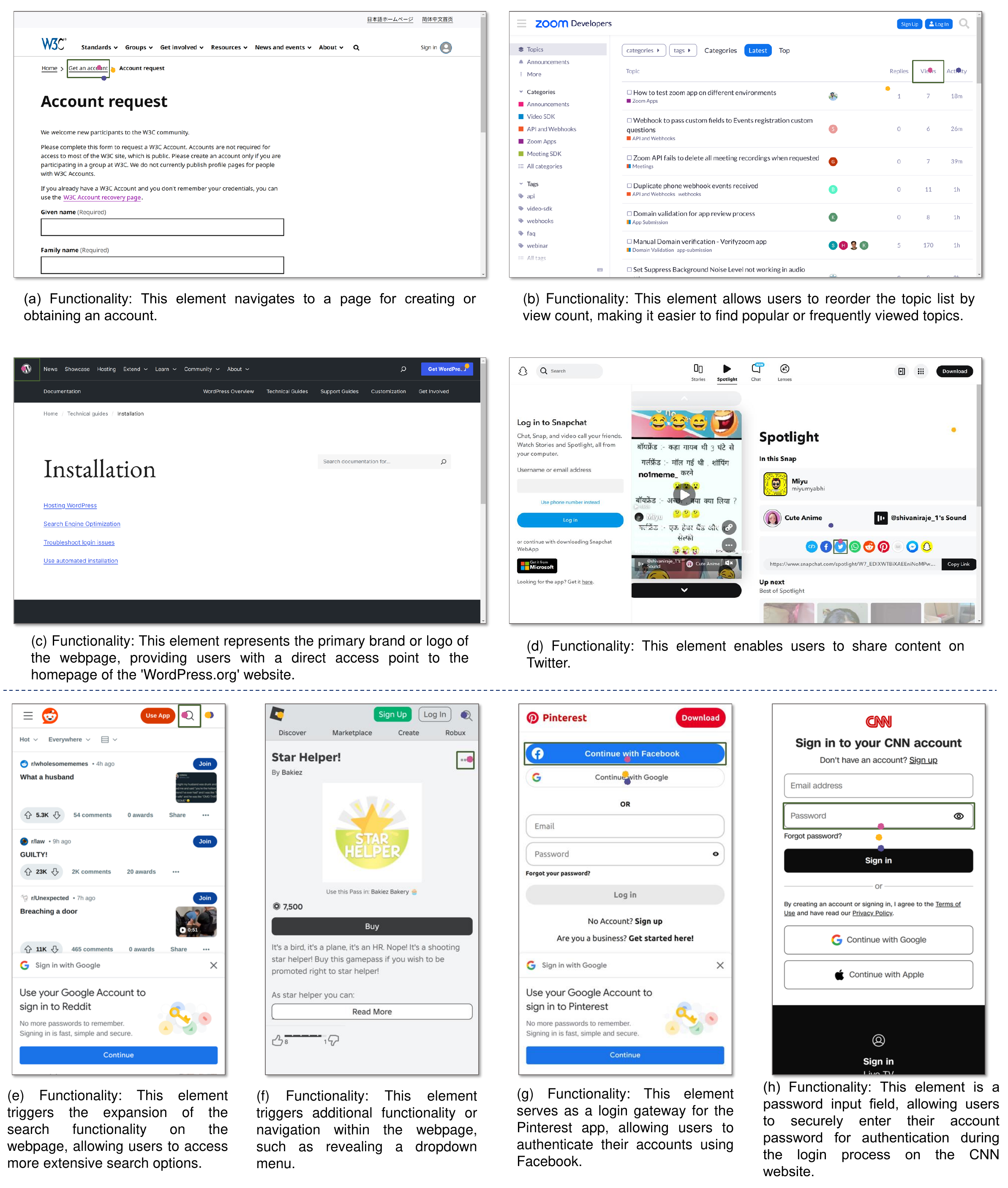}
    \caption{\textbf{Visualization of the successful functionality grounding examples for ours-625k.} The ground truth bounding boxes, ours-625k predictions, ours-125k predictions, and ours-25k predictions are drawn in \textcolor[rgb]{0, 0.6, 0.13}{green}, \textcolor[rgb]{1.0, 0.0, 1.0}{pink}, \textcolor[rgb]{0.0, 0.4, 0.8}{blue}, and \textcolor[rgb]{1.0, 0.65, 0.0}{orange}, respectively.}
    \label{fig:funcpred scaling success}
\end{figure*}

\noindent\textbf{Successful cases} Fig.~\ref{fig:funcpred scaling success} demonstrates several examples of the grounding results from Qwen-VL trained with the 25k, 125k, and 702k AutoGUI data. The model trained with the 702k data (ours-702k) exhibits more accurate functionality grounding performance. For instance, Fig.~\ref{fig:funcpred scaling success} (a) shows that ours-702k predicts the point right on the target (The `Get an account' button) while the other two models slightly miss the target. Case (c) shows that ours-702k correctly understands the functional intent to locate the WordPress logo, in contrast to the other models, which incorrectly focus on the text `Get WordPress'. Additionally, case (f) illustrates that ours-702k successfully locates the three-dot menu icon, aligning with the intent to expand a dropdown menu. These results suggest that increasing the AutoGUI training data enhances the model's ability to understand complex functional intents and to recognize diverse iconic elements accurately.

\begin{figure*}[th]
    \centering
    \includegraphics[width=1.0\linewidth]{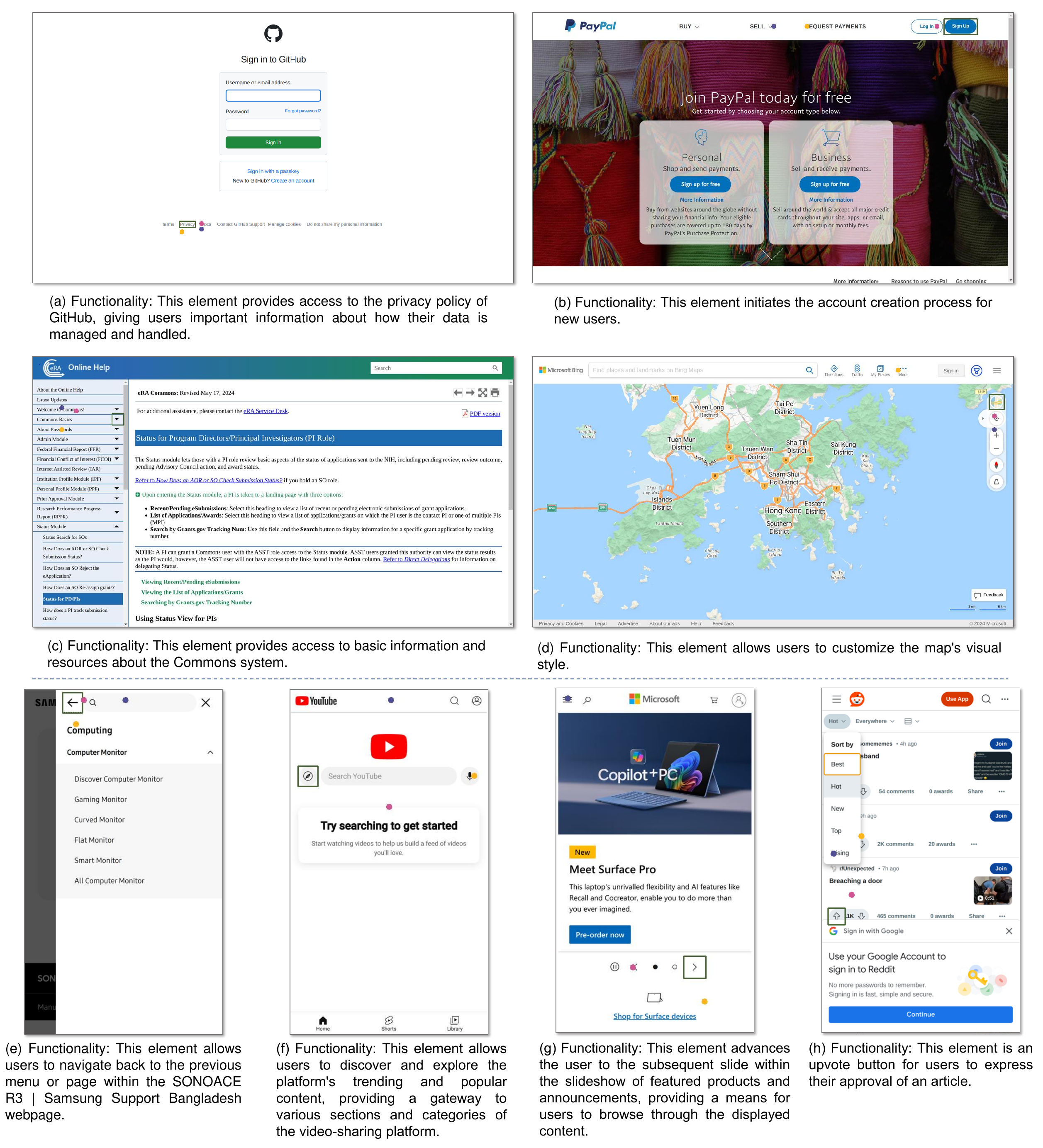}
    \caption{\textbf{Visualization of failure examples in the scaling experiments.} The ground truth bounding boxes, ours-625k predictions, ours-125k predictions, and ours-25k predictions are drawn in \textcolor[rgb]{0, 0.6, 0.13}{green}, \textcolor[rgb]{1.0, 0.0, 1.0}{pink}, \textcolor[rgb]{0.0, 0.4, 0.8}{blue}, and \textcolor[rgb]{1.0, 0.65, 0.0}{orange}, respectively.}
    \label{fig:funcpred scaling failure}
\end{figure*}

\noindent\textbf{Failure cases} To explore the limitations of our model, we analyze several failure cases across the scaling experiments, as shown in Fig.~\ref{fig:funcpred scaling failure}. The primary failure cases comprise (1) Difficulty in accurately locating very small target elements, as illustrated by the tiny ‘Policy’ button in case (a); (2) Misunderstanding functional intents, as shown in case (b) where the three models fail to locate the element for account creation and case (g) where ours-702k mistakenly focuses on navigating to previous content instead of subsequent content; (3) Challenges in recognizing abstract iconic elements, as seen with the map style icon in case (d) and the compass icon in case (f).

Despite these challenges, the enhanced performance observed with ours-702k supports the potential of the AutoGUI pipeline to further improve functionality grounding. The successful cases underscore that increasing the size of the training dataset not only boosts the model’s ability to interpret functional intents but also its capability to process a variety of textual and iconic elements effectively.

\subsubsection{Case Analysis on MoTIF Test Split}
\begin{figure*}[h]

    \vspace{-12mm}
    \centering
    \includegraphics[width=\textwidth]{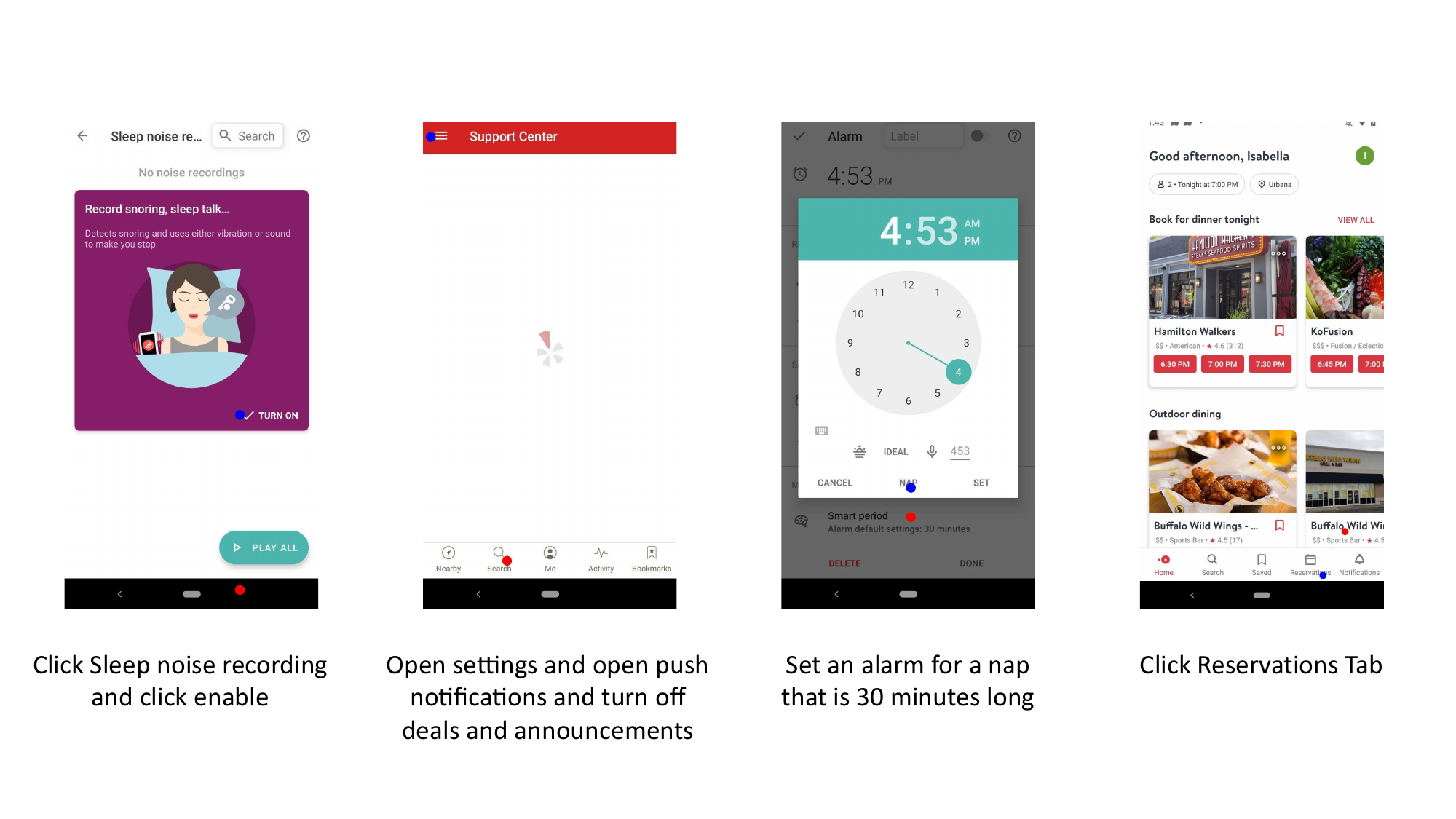}
    \vspace{-2mm}
    \caption{Evaluation results of the model trained on \textcolor{blue}{625k} (blue dot) and \textcolor{red}{125k} (red dot).}
    \vspace{-3mm}
    \label{fig: motif case}
\end{figure*}
\begin{figure*}[h]
    \vspace{-12mm}
    \centering
    \includegraphics[width=\textwidth]{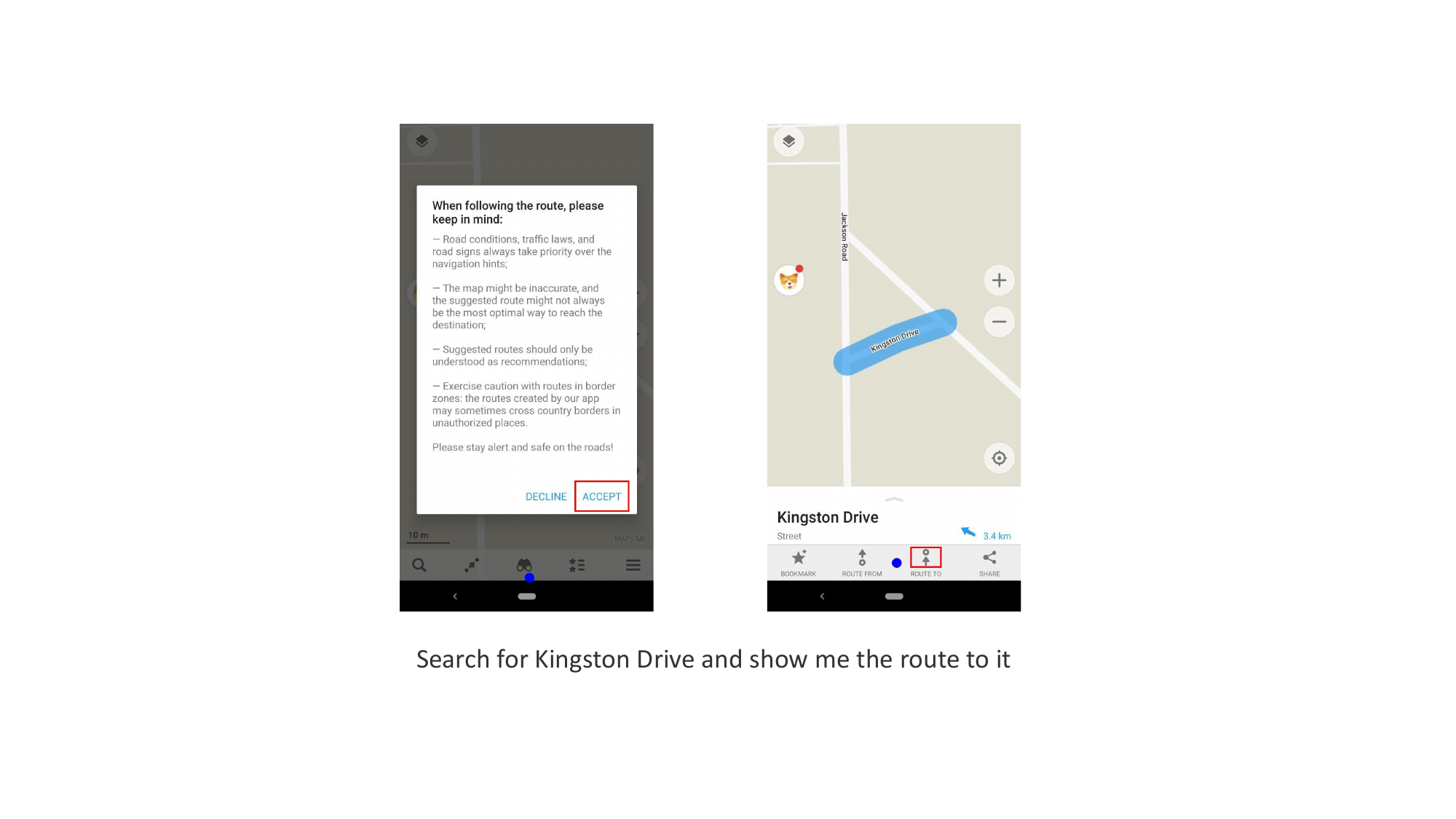}
    \vspace{-2mm}
    \caption{Bad cases on MoTIF.}
    \vspace{-3mm}
    \label{fig: motif bad case}
\end{figure*}

We evaluate the instruction following ability on MoTIF dataset. Our analysis focuses on two aspects: (1) what improvements our model can achieve with the scaling of our functionality dataset (Fig.~\ref{fig: motif case}); and (2)  in which scenarios our model still fails to achieve correct grounding (Fig.~\ref{fig: motif bad case}).

Fig.~\ref{fig: motif case} shows that the model can more accurately understand the action instruction and make meaningful localization as scaling improves from \textcolor{red}{125k} to \textcolor{blue}{702k}. For instance, when the objective is to \textit{click sleep noise recording and click enable}, the model can comprehend the semantics of this global objective and identify \textit{turn on}. Additionally, the model can mitigate localization errors, such as the \textcolor{blue}{702k} being more accurately positioned on the target element (e.g., the icon of \textit{reservation}) than the \textcolor{red}{125k}.
However, MoTIF still struggles with certain tasks. For example, as shown Fig.~\ref{fig: motif bad case}, it has difficulty with localization in fine-grained steps for the instruction \textit{search for Kingston Drive and show me the route to it}. It can be seen that the model does not effectively understand situations involving widget pop-ups (e.g., protocol and advertisement). This may be attributed to the weak semantic connection between pop-ups and the instruction. Furthermore, the model still falls short in precise localization. Enriching the dataset further could alleviate this issue.

\begin{figure*}[t]
    \centering
    \includegraphics[width=0.95\linewidth]{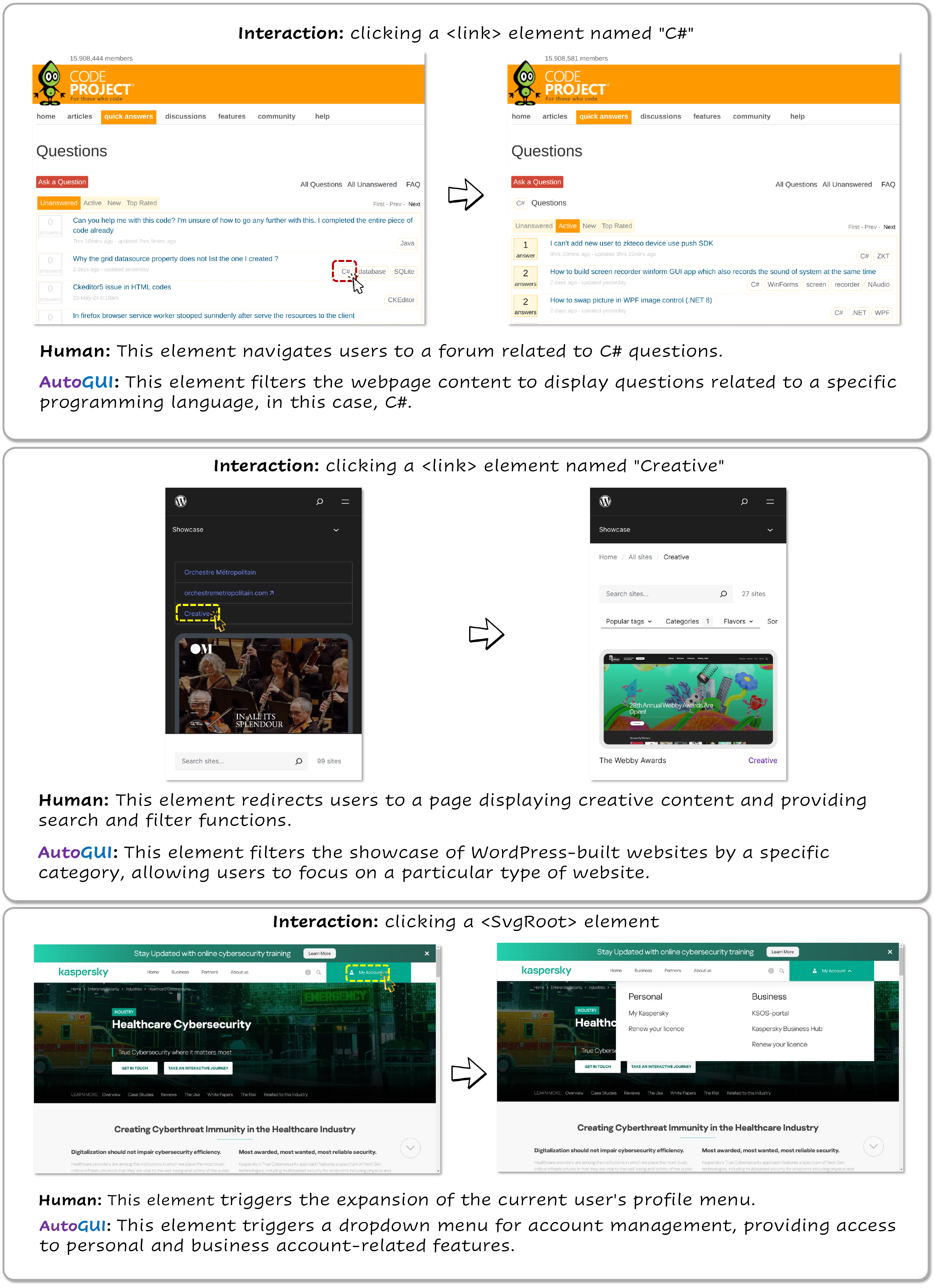}
    \caption{Comparing the annotations generated by a trained human annotator and the proposed AutoGUI pipeline. We can see that AutoGUI annotations are more detailed and clear than those by the human annotator.}
    \label{fig: autogui vs human}
\end{figure*}

\subsection{Potential Societal Impact}
The potential societal impacts of the proposed AutoGUI can be considered across various dimensions:

\noindent\textbf{Accessibility Enhancements} VLMs trained with the AutoGUI data obtain stronger UI grounding capabilities, thereby possessing the potential to act as UI agents. By enabling context-aware understanding of UI functionalities, the VLMs can help users locate elements on complex UIs, significantly improving accessibility features in software. This could lead to the development of applications that are more intuitive for users with disabilities, such as those requiring screen readers or other assistive technologies.

\noindent\textbf{Research Impact}: By reducing the labor and time required for annotating UI data via the AutoGUI, the industry and academia could lower costs to easily build UI agents. This could also shift labor demands towards more creative and strategic roles rather than repetitive annotation tasks.

\noindent\textbf{Privacy and Security Concerns}: Although we employ precautions of eliminating samples related to sensitive UI elements (e.g., avoid interacting with elements modifying the Internet and use only popular public websites without exposing privacy), corner cases still exist on the vast Internet. UI data involving either content modification or personal information are hard to discern as UI designs are distinct and no universal detection rules exist. Therefore, it is essential for cyber-security research to consider the potential leakage of personal information in the collected data and devise preemptive protective approaches.

\noindent\textbf{Potential for Bias and Fairness}: The bias of the LLMs used in the AutoGUI annotation pipeline is probably reflected in the collected data, leading to a trained UI-VLM that inherits the bias. Therefore, mitigating bias in the LLM's annotations will be important for developing fair VLM agents that align with the values of users from diverse cultures.

\end{document}